\documentclass[runningheads]{llncs}
\usepackage{graphicx}
\usepackage{amsmath,amssymb} 
\usepackage{hyperref}
\usepackage{color}
\usepackage{times}
\usepackage{epsfig}
\usepackage{graphicx}
\usepackage{amsmath}
\usepackage{amssymb}
\usepackage{graphics}
\usepackage{graphicx}
\usepackage[]{graphicx} 
\usepackage{epsfig} 
\usepackage{subfigure}
\usepackage{algorithm}
\usepackage{algorithmic}
\usepackage{stmaryrd}
\usepackage[mathscr]{eucal}
\usepackage{lineno}
\usepackage{color}
\usepackage{filecontents}
\usepackage{subfigure}
\usepackage{multirow}
\usepackage{amsmath}
\usepackage{amssymb}
\usepackage{filecontents}
\usepackage{verbatim} 
\usepackage{tabularx}
\usepackage{lineno}
\usepackage{setspace}
\usepackage{multirow}
\usepackage{textcomp,booktabs}

\usepackage[utf8]{inputenc}

\def\0{{\bf 0}}
\def\1{{\bf 1}}

\def\eg{\emph{e.g. }}
\def\ie{\emph{i.e. }}

\def\etal{{\em et al.\/}\,}

\graphicspath{{./figures/}}   
\usepackage{color}\usepackage[width=122mm,left=12mm,paperwidth=146mm,height=193mm,top=12mm,paperheight=217mm]{geometry}
\begin{document}
\pagestyle{headings}
\mainmatter
\title{Photo Aesthetics Ranking Network with \\ Attributes and Content Adaptation} 

\titlerunning{Photo Aesthetics Ranking Network with  Attributes and Content Adaptation}

\authorrunning{Shu Kong, Xiaohui Shen, Zhe Lin, Radomir Mech, Charless Fowlkes}

\author{Shu Kong$^\star$, Xiaohui Shen$^*$, Zhe Lin$^*$, Radomir Mech$^*$, Charless Fowlkes$^\star$}
\institute{ $^\star$UC Irvine \ \ \ \ \  \ \ \  \ {\tt \{skong2,fowlkes\}@ics.uci.edu}\\
        $^*$Adobe \ \ \ \ \  \ \ \ \ \ \ \ \ \ {\tt \{xshen,zlin,rmech\}@adobe.com} \\
  \href{http://www.ics.uci.edu/~skong2/aesthetics.html}{project webpage}  }

\maketitle

\begin{abstract}
Real-world applications could benefit from the ability to automatically generate
a fine-grained ranking of photo aesthetics. However, previous methods for image
aesthetics analysis have primarily focused on the coarse, binary categorization of
images into high- or low-aesthetic categories.  In this work, we
propose to learn a deep convolutional neural network to rank photo
aesthetics in which the relative ranking of photo aesthetics are directly
modeled in the loss function. Our model incorporates joint learning of
meaningful photographic attributes and image content information which can help
regularize the complicated photo aesthetics rating problem.

To train and analyze this model, we have assembled a new aesthetics and attributes database (AADB)
which contains aesthetic scores and meaningful attributes assigned to each
image by multiple human raters. Anonymized rater identities are recorded across
images allowing us to exploit intra-rater consistency using a novel sampling
strategy when computing the ranking loss of training image pairs.  We show the
proposed sampling strategy is very effective and robust in face of subjective judgement of
image aesthetics by individuals with different aesthetic tastes. Experiments
demonstrate that our unified model can generate aesthetic rankings that are
more consistent with human ratings. To further validate our model, we show that
by simply thresholding the estimated aesthetic scores, we are able to achieve
state-or-the-art classification performance on the existing AVA dataset benchmark.
\end{abstract}

\section{Introduction}
\label{sec:intro}

\begin{figure}
\centering
   \includegraphics[width=0.75\linewidth]{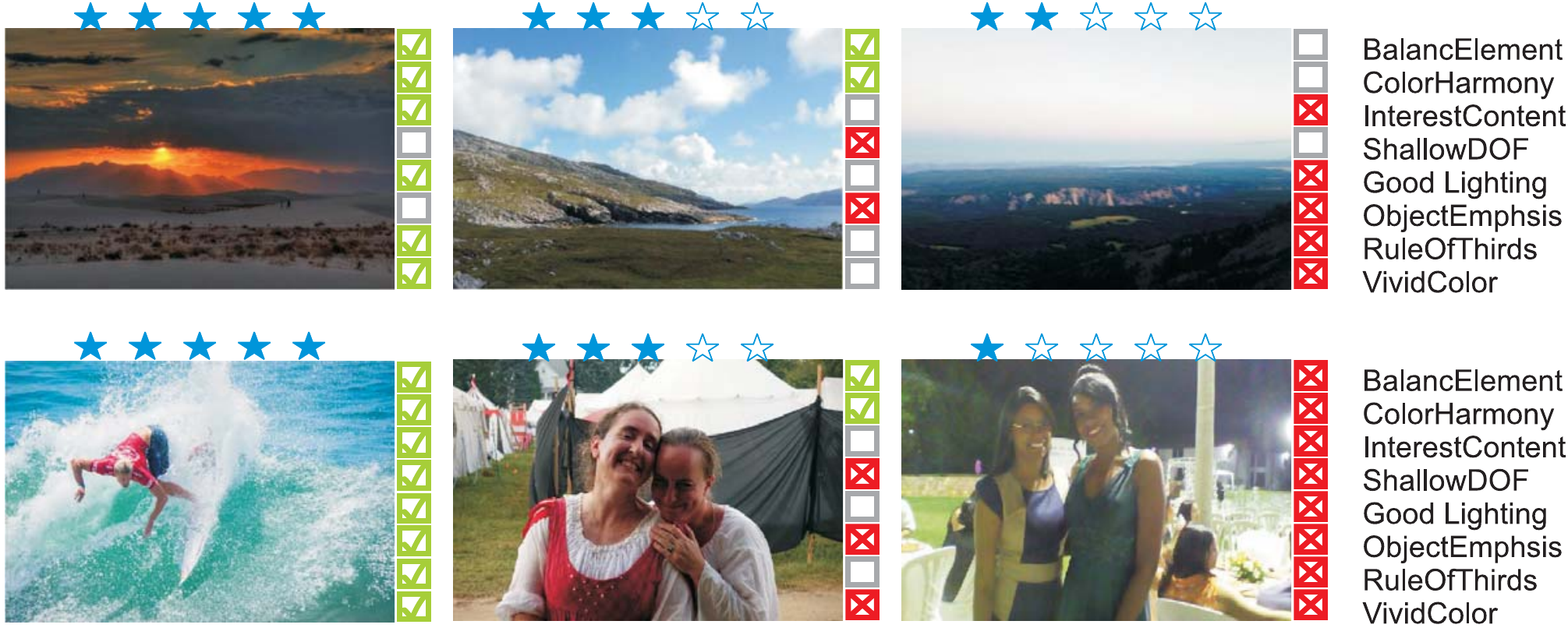}
   \caption{Classification-based methods for aesthetic analysis can distinguish
   high- and low-quality images shown in the leftmost and rightmost columns, but fail to
   provide useful insights about borderline images displayed in the middle column.
   This observation motivates us to consider rating and ranking images w.r.t
   aesthetics rather than simply assigning binary labels. We observe that the
   contribution of particular photographic attributes to making an image
   aesthetically pleasing depends on the thematic content (shown in different
   rows), so we develop a model for rating that incorporates joint
   attributes and content.  The attributes and ratings of aesthetics on a scale 1 to 5
   are predicted by our model (displayed on top and right of each image, respectively).
   }
\vspace{-7mm}
\label{fig:motivation}
\end{figure}

Automatically assessing image aesthetics is increasingly important for a variety of applications~\cite{marchesotti2014discovering,xin2015iccv},
including personal photo album management, automatic photo editing, and image retrieval.
While judging image aesthetics is a subjective task,
it has been an area of active study in recent years and substantial progress has been made
in identifying and quantifying those image features that are predictive of favorable aesthetic judgements by most individuals
~\cite{marchesotti2014discovering,xin2015iccv,lu2014rapid,luo2011content,datta2006studying}.

Early works formulate aesthetic analysis as a classification or a regression
problem of mapping images to aesthetic ratings provided by human
raters~\cite{datta2006studying,ke2006design,luo2011content,dhar2011high,nishiyama2011aesthetic}.
Some approaches have focused on designing hand-crafted features that encapsulate standard
photographic practice and rules of visual design, utilizing both low-level statistics (\eg color histogram and wavelet analysis)
and high-level cues based on traditional photographic rules (\eg region composition and rule of thirds).
Others have adopted generic image content features, which are originally designed
for recognition (\eg SIFT~\cite{lowe2004distinctive} and Fisher
Vector~\cite{perronnin2010improving,perronnin2007fisher}),
that have been found to outperform methods using rule-based features~\cite{marchesotti2011assessing}.
With the advance of deep Convolutional Neural Network (CNN)~\cite{krizhevsky2012imagenet},
recent works propose to train end-to-end models for image aesthetics classification~\cite{kang2014convolutional,lu2014rapid,xin2015iccv},
yielding state-of-the-art performance on a recently released Aesthetics Visual Analysis dataset (AVA)~\cite{murray2012ava}.


Despite notable recent progress towards computational image aesthetics classification (\eg~\cite{lu2014rapid,marchesotti2014discovering,xin2015iccv}),
judging image aesthetics is still a subjective task,
and it is difficult to learn a universal scoring mechanism for various kinds of images.
For example,
as demonstrated in Fig.~\ref{fig:motivation},
images with obviously visible high- or low- aesthetics are relatively easy to classify,
but existing methods cannot generate reliable labels for borderline images.
Therefore,
instead of formulating image aesthetics analysis as an overall binary classification or regression problem,
we argue that it is far more practical and useful to predict relative aesthetic
rankings among images with similar visual content along with generating richer
descriptions in terms of aesthetic attributes~\cite{Geng2011WebSearch,san2012leveraging}.

To this end,
we propose to train a model through a Siamese network~\cite{chopra2005learning} that takes a pair of images as input and directly predicts relative ranking of their aesthetics in addition to their overall aesthetic scores.
Such a structure allows us to deploy different sampling strategies of image pairs and leverage auxiliary side-information to regularize the training,
including aesthetic attributes~\cite{dhar2011high,lu2014rapid,marchesotti2014discovering} and photo content~\cite{luo2011content,murray2012ava,xin2015TMM}.
For example,
Fig.~\ref{fig:motivation} demonstrates that photos with different contents convey different attributes to make them aesthetically pleasing.
While such side information has been individually adopted to improve aesthetics classification~\cite{lu2014rapid,marchesotti2014discovering},
it remains one open problem to systematically incorporate all the needed components in a single end-to-end framework with fine-grained aesthetics ranking.
Our model and training procedure naturally incorporates both attributes and content information by
sampling image pairs with similar content to learn the specific relations of attributes and aesthetics for different content sub-categories.
As we show, this results in more comparable and consistent aesthetics estimation results.


Moreover,
as individuals have different aesthetics tastes,
we argue that it is important to compare ratings assigned by an individual across multiple images in order to provide a more consistent training signal.
To this end,
we have collected and will publicly release a new dataset in which each image is associated with a detailed score distribution,
meaningful attributes annotation and (anonymized) raters' identities.
We refer to this dataset as the ``Aesthetics with Attributes Database'', or AADB for short.
AADB not only contains a much more balanced distribution of professional and consumer photos
and a more diverse range of photo qualities than available in the exiting AVA dataset,
but also identifies ratings made by the same users across multiple images.
This enables us to develop novel sampling strategies for training our model which focuses on relative rankings by individual raters.
Interestingly,
this rater-related information also enables us to compare the trained model to each individual's rating results
by computing the ranking correlation over test images rated by that individual.
Our experiments show the effectiveness of the proposed model in rating image aesthetics compared to human individuals.
We also show that, by simply thresholding rated aesthetics scores,
our model achieves state-of-the-art classification performance on the AVA dataset,
even though we do not explicitly train or tune the model for the aesthetic classification task.

In summary, our main contributions are three-fold:
\begin{enumerate}
  \item We release a new dataset containing not only score distributions, but also informative attributes and anonymized rater identities.
        These annotations enable us to study the use of individuals' aesthetics ratings for training our model
        and analyze how the trained model performs compared to individual human raters.
  \item We propose a new CNN architecture that unifies aesthetics attributes and photo content for image aesthetics rating
    and achieves state-of-the-art performance on existing aesthetics classification benchmark.
  \item We propose a novel sampling strategy that utilizes mixed within- and cross-rater image pairs for training models.
        We show this strategy, in combination with pair-wise ranking loss, substantially improves the performance \emph{w.r.t.} the ranking correlation metric.
\end{enumerate}

\section{Related Work}
\label{sec:relatedWork}

\ \ \ \ \  \textbf{CNN for aesthetics classification: }
In~\cite{lu2014rapid,kang2014convolutional,xin2015iccv},
CNN-based methods are proposed for classifying images into high- or low- aesthetic categories.
The authors also show that using patches from the original high-resolution images largely improves the performance.
In contrast,
our approach formulates aesthetic prediction as a combined regression and ranking problem.
Rather than using patches, our architecture warps the whole input image in order to minimize
the overall network size and computational workload while retaining compositional elements in the image,
\eg rule of thirds, which are lost in patch-based approaches.

\textbf{Attribute-adaptive models: }
Some recent works have explored the use of high-level describable
attributes~\cite{dhar2011high,marchesotti2014discovering,lu2014rapid} for image
aesthetics classification.
In early work, these attributes were modeled using hand-crafted features~\cite{dhar2011high}.
This introduces some intrinsic problems, since
(1) engineering features that capture high-level semantic attributes is a difficult task,
and (2) the choice of describable attributes may ignore some aspects of the image which
are relevant to the overall image aesthetics.
For these reasons,
Marchesotti \etal propose to automatically select a large number of useful attributes based on textual comments from raters~\cite{marchesotti2013learning}
and model these attributes using generic features~\cite{marchesotti2011assessing}.
Despite good performance,
many of the discovered textual attributes (\eg $so\_cute$, $those\_eyes$,
$so\_close$, $very\_busy$, $nice\_try$) do not correspond to well defined visual characteristics which hinders their
detectability and utility in applications.
Perhaps the closest work to our approach is that of Lu \etal, who propose to
learn several meaningful style attributes~\cite{lu2014rapid} in a CNN
framework and use the hidden features to regularize aesthetics classification network training.

\textbf{Content-adaptive models: }
To make use of image content information such as scene categories or choice of
photographic subject, Luo \etal propose to segment regions and extract visual
features based on the categorization of photo content~\cite{luo2011content}.
Other work, such as~\cite{murray2012ava,xin2015TMM}, has also demonstrated that
image content is useful for aesthetics analysis. However, it has been assumed
that the category labels are provided both during  training and testing.
To our knowledge,
there is only one paper~\cite{murray2012learning} that attempts to jointly predict content semantics and aesthetics labels.
In~\cite{murray2012learning},
Murray \etal propose to rank images w.r.t aesthetics in a three-way classification problem (high-, medium- and low-aesthetics quality).
However,
their work has some limitations because (1) deciding the thresholds between nearby classes is non-trivial,
and (2) the final classification model outputs a hard label which is less useful than a continuous rating.

Our work is thus unique in
presenting a unified framework that is trained by jointly incorporating the photo content,
the meaningful attributes and the aesthetics rating in a single CNN model.
We train a category-level classification
layer on top of our aesthetics rating network to generate soft weights of category labels,
which are used to combine scores predicted by multiple content-adaptive branches.
This allows category-specific subnets to complement each other in rating image aesthetics
with shared visual content information while efficiently re-using front-end
feature computations. While our primary focus is on aesthetic rating prediction,
we believe that the content and attribute predictions (as displayed on the
right side of images in Fig.~\ref{fig:motivation}) represented in hidden layers
of our architecture could also be surfaced for use in other applications such
as automatic image enhancement and image retrieval.

\section{Aesthetics and Attributes Database}
\label{ssec:datasetCollection}

\begin{table}[t]
\small
\centering
\begin{tabular}{|l|c|c|c|c|}
\hline
            &	AADB&		AVA~\cite{murray2012ava}    & PN~\cite{datta2006studying}     & CUHKPQ\cite{ke2006design,luo2008photo} \\
\hline\hline
Rater's ID         &	\textcolor[rgb]{0.00,1.00,0.25}{Y}  &	\textcolor[rgb]{1.00,0.00,0.00}{N}     &	\textcolor[rgb]{1.00,0.00,0.00}{N}     &	\textcolor[rgb]{1.00,0.00,0.00}{N} 	\\
All Real Photo     &	\textcolor[rgb]{0.00,1.00,0.25}{Y}  &	\textcolor[rgb]{1.00,0.00,0.00}{N}     &	\textcolor[rgb]{0.00,1.00,0.25}{Y}     &	\textcolor[rgb]{0.00,1.00,0.25}{Y}	\\
Attribute Label        &	\textcolor[rgb]{0.00,1.00,0.25}{Y}  &	\textcolor[rgb]{0.00,1.00,0.25}{Y}     &	\textcolor[rgb]{1.00,0.00,0.00}{N}     &	\textcolor[rgb]{1.00,0.00,0.00}{N}	\\
Score Dist.        &	\textcolor[rgb]{0.00,1.00,0.25}{Y}  &	\textcolor[rgb]{0.00,1.00,0.25}{Y}     &	\textcolor[rgb]{0.00,1.00,0.25}{Y}     &	\textcolor[rgb]{1.00,0.00,0.00}{N}	\\
\hline
\end{tabular}
\caption{Comparison of the properties of current image aesthetics datasets.  In
addition to score distribution and meaningful style attributes, AADB also
tracks raters' identities across images which we exploit in training to improve
aesthetic ranking models.}
\label{tab:datasetComparison}
\vspace{-3mm}
\end{table}

\begin{figure}[t]
\centering
   \includegraphics[width=0.90\linewidth]{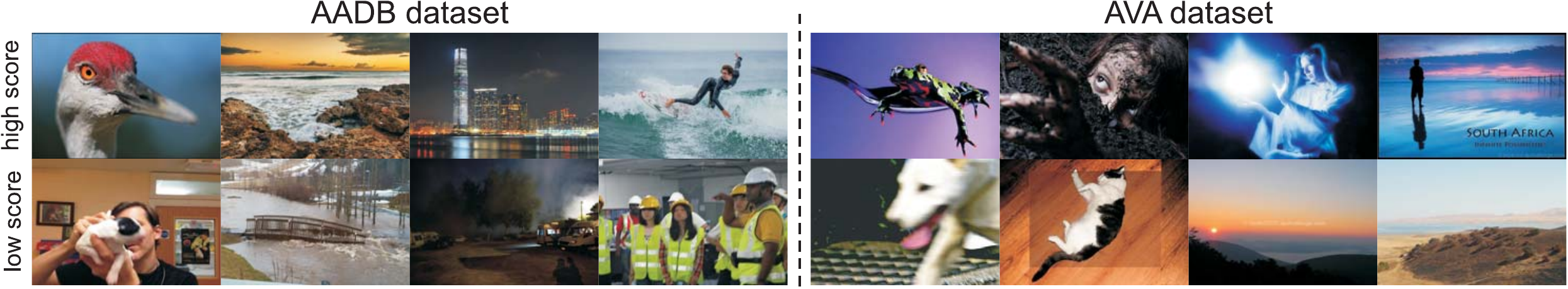}
\vspace*{-1mm}
   \caption{Our AADB dataset consists of a wide variety of photographic imagery
   of real scenes collected from Flickr. This differs from AVA which contains
   significant numbers of professional images that have been highly
   manipulated, overlayed with advertising text, etc.}
\label{fig:examplesTwoDB}
\vspace{-1mm}
\end{figure}

To collect a large and varied set of photographic images, we download images
from the Flickr website\footnote{www.flickr.com}
which carry a Creative Commons
license and manually curate the data set to remove non-photographic images
(\eg cartoons, drawings, paintings, ads images, adult-content images, etc.).
We have five different workers then independently annotate each image with an overall
aesthetic score and a fixed set of eleven meaningful attributes using Amazon
Mechanical Turk (AMT)\footnote{www.mturk.com}.
The AMT raters work on batches,
each of which contains ten images.
For each image,
we average the ratings of five raters as the ground-truth aesthetic score.
The number of images rated by a
particular worker follows long tail distribution,
as shown later in Fig.~\ref{fig:workerPerformanceDistribution} in the experiment.

After consulting professional photographers, we selected eleven attributes that
are closely related to image aesthetic judgements: $interesting\_content$, $object\_emphasis$,
$good\_lighting$, $color\_harmony$, $vivid\_color$, $shallow\_depth\_of\_field$,
$motion\_blur$, $rule\_of\_thirds$, $balancing\_element$, $repetition$, and $symmetry$.
These attributes span traditional photographic principals of color, lighting,
focus and composition, and provide a natural vocabulary for use in applications,
such as auto photo editing and image retrieval.  The final AADB dataset
contains  10,000 images in total, each of which have aesthetic quality
ratings and attribute assignments provided by five different individual raters.
Aggregating multiple raters allows us to assign a confidence score to each
attribute, unlike, e.g., AVA where attributes are binary.  Similar to previous
rating datasets~\cite{murray2012ava}, we find that average ratings are well fit
by a Gaussian distribution.  For evaluation purposes, we randomly split the
dataset into validation (500), testing (1,000) and training sets (the rest).
The supplemental material provides additional details about dataset collection
and statistics of the resulting data.

Table~\ref{tab:datasetComparison} provides a summary comparison of AADB to other related public databases for image aesthetics analysis.
Except for our AADB and the existing AVA dataset, many existing datasets have
two intrinsic problems (as discussed in~\cite{murray2012ava}),
(1) they do not provide full score distributions or style attribute annotation,
and (2) images in these datasets are either biased or consist of examples which
are particularly easy for binary aesthetics classification. Datasets such as
CUHKPQ~\cite{ke2006design,luo2008photo} only
provide binary labels (low or high aesthetics) which cannot easily be used for
rating prediction. A key difference between our dataset and AVA is that many
images in AVA are heavily edited or synthetic (see
Fig.~\ref{fig:examplesTwoDB}) while AADB contains a much more balanced distribution of professional and consumer photos.
More importantly, AVA does not provide any way to identify ratings provided by
the same individual for multiple images. We report results of experiments, showing that
rater identity on training data provides useful side information for training
improved aesthetic predictors.

\vspace{2mm}
\noindent \textbf{Consistency Analysis of the Annotation:}
One concern is that the annotations provided by five AMT workers for each
image may not be reliable given the subjective nature of the task.
Therefore, we conduct consistency analysis on the annotations.
Since the same five workers annotate a batch of ten images,
we study the consistency at batch level.  We use Spearman's rank correlation
$\rho$ between pairs of workers to measure consistency within a batch and
estimate $p$-values to evaluate statistical significance of the correlation
relative to a null hypothesis of uncorrelated responses.  We use the
Benjamini-Hochberg procedure to control the false discovery rate (FDR) for
multiple comparisons~\cite{benjamini2001control}.  At an FDR level of $0.05$,
we find $98.45\%$ batches have significant agreement among raters.  This shows
that the annotations are reliable for scientific research.  Further consistency
analysis of the dataset can be found in the supplementary material.

\section{Fusing Attributes and Content for Aesthetics Ranking}
\label{sec:ourModels}

Inspired by~\cite{simonyan2014very,xin2015iccv},
we start by fine-tuning AlexNet~\cite{krizhevsky2012imagenet} using regression loss to predict aesthetic ratings.
We then fine-tune a Siamese network~\cite{chopra2005learning} which takes image pairs as input and is
trained with a joint Euclidean and ranking loss (Section~\ref{ssec:rankloss}).
We then append attribute (Section~\ref{ssec:attributeNet})
and content category classification layers (Section~\ref{ssec:contentNet})
and perform joint optimization.


\subsection{Regression Network for Aesthetics Rating}
\label{ssec:regNet}
The network used in our image aesthetics rating is fine-tuned from AlexNet~\cite{krizhevsky2012imagenet} which is used for image classification.
Since our initial model predicts a continuous aesthetic score other than category labels,
we replace the softmax loss with the Euclidean loss given by
$loss_{reg} =  \frac{1}{2N} \sum_{i=1}^{N} \Vert \hat y_i - y_i \Vert_2^2$,
where $y_i$ is the average ground-truth rating for image-$i$,
and  $\hat y_i$ is the estimated score by the CNN model.
Throughout our work, we re-scale all the ground-truth ratings to be in the
range of $[0,1]$ when preparing the data.
Consistent with observations in~\cite{xin2015iccv}, we find that fine-tuning
the pre-trained AlexNet~\cite{krizhevsky2012imagenet} model
performs better than that training the network from scratch.

\begin{figure}[t]
\centering
   \includegraphics[width=0.9\linewidth]{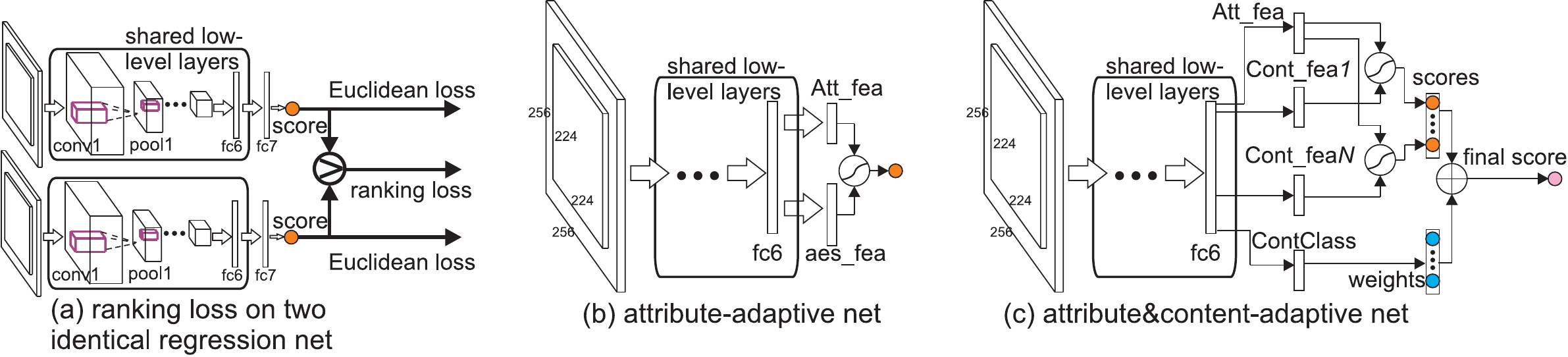}
   \caption{Architectures for our different models. All models utilize the
   AlexNet front-end architecture which we augment by (a) replacing the top
   softmax layer with a regression net and adopting ranking loss in addition to
   Euclidean loss for training, (b) adding an attribute predictor branch which
   is then fused with the aesthetic branch to produce a final attribute-adapted
   rating and (c) incorporating image content scores that act as weights to
   gate the combination of predictions from multiple content-specific branches.}
\vspace{-2mm}
\label{fig:net_summary}
\end{figure}


\subsection{Pairwise Training and Sampling Strategies}
\label{ssec:rankloss}
A model trained solely to minimize the Euclidean loss may still make mistakes
in the relative rankings of images that have similar average aesthetic scores.
However, more accurate fine-grained ranking of image aesthetics is quite
important in applications (\eg in automating photo album
management~\cite{cui2007easyalbum}).
Therefore,
based on the Siamese network~\cite{chopra2005learning},
we adopt a pairwise ranking loss to explicitly exploit relative rankings of image pairs
available in the AADB data (see Fig.~\ref{fig:net_summary} (a)).  The ranking loss is
given by:
\begin{equation}
\small
\begin{split}
loss_{rank} = & \frac{1}{2N} \sum_{i,j} \max \big(0, \alpha - \delta(y_i \geq y_j)(\hat y_i  - \hat y_j ) \big) \\
\end{split}
\label{eq:rankloss}
\end{equation}
where $\delta(y_i \geq y_j)= \left \{
 \begin{aligned}
        1, \qquad \text{if } y_i \ge y_j \\
        -1, \qquad \text{if } y_i < y_j
       \end{aligned}
       \right.
$,
and $\alpha$ is a specified margin parameter. By adjusting this margin and
the sampling of image pairs, we can avoid the need to sample triplets as done
in previous work on learning domain-specific similarity
metrics~\cite{chopra2005learning,wang2014learning,schroff2015facenet}.
Note that the regression alone focuses the capacity of the network on predicting the commonly occurring range of
scores,
while ranking penalizes mistakes for extreme scores more heavily.

In order to anchor the scores output by the ranker to the same scale as
user ratings, we utilize a joint loss function that includes both ranking
and regression:
\begin{equation}
\small
loss_{reg+rank} = loss_{reg}+ \omega_r loss_{rank},
\label{eq:regRankModel}
\end{equation}
where the parameter $\omega_r$ controls the relative importance of the ranking loss and is
set based on validation data. The network structure is shown in Fig.~\ref{fig:net_summary} (a).

Such a structure allows us to utilize different pair-sampling strategies to narrow the scope of learning and provide more consistent training.
In our work,
we investigate two strategies for selecting pairs of
images used in computing the ranking loss.
First, we can bias sampling towards
pairs of images with a relatively large difference in their average aesthetic scores.
For these pairs, the ground-truth rank order is likely to be stable (agreed
upon by most raters).  Second, as we have raters' identities across images, we
can sample image pairs that have been scored by the same individual.
While different raters may have different aesthetics tastes which erode differences
in the average aesthetic score, we expect a given individual should have more
consistent aesthetic judgements across multiple images.  We show the empirical
effectiveness of these sampling strategies in Section~\ref{sec:exp}.

\subsection{Attribute-Adaptive Model}
\label{ssec:attributeNet}
Previous work on aesthetic prediction has investigated the use of attribute
labels as input features for aesthetics classification (\eg \cite{dhar2011high}).
Rather than independently training attribute classifiers,
we propose to include
additional activation layers in our ranking network that are trained to encode
informative attributes.
We accomplish this by including an additional term in
the loss function that encourages the appropriate attribute activations.  In
practice, annotating attributes for each training image is expensive and time
consuming.  This approach has the advantage that it can be used even when
only a subset of training data comes with attribute annotations.  Our approach
is inspired by~\cite{lu2014rapid} which also integrates attribute classifiers,
but differs in that the attribute-related layer shares the same front-end
feature extraction with the aesthetic score predictor (see
Fig.~\ref{fig:net_summary}(b)). The attribute prediction task can thus be
viewed as a source of side-information or ``deep supervision''
\cite{zhuowentuAISTATSpaper} that serves to regularize the weights learned
during training even though it is not part of the test-time prediction,
though could be enabled when needed.

We add an attribute prediction branch on top of the second fully-connected
layer in the aesthetics-rating network described previously. The attribute
predictions from this layer are concatenated with the base model to predict
the final aesthetic score.  When attribute annotations are available, we
utilize a $K$-way softmax loss or Euclidean loss, denoted by $loss_{att}$, for the attribute
activations and combine it with the rating and ranking losses
\begin{equation}
\begin{split}
loss = &  loss_{reg}+ \omega_r loss_{rank}   + \omega_{a} loss_{att}
\end{split}
\label{eq:regRankAttModel}
\end{equation}
where $\omega_a$ controls relative importance of attribute fine-tuning.
If we do not have enough data with attribute annotations, we can freeze the
attribute layer and only fine-tune through the other half of the concatenation
layer.

\subsection{Content-Adaptive Model}
\label{ssec:contentNet}
The importance of particular photographic attributes depends strongly on image
content~\cite{luo2011content}.  For example, as demonstrated by Fig.~\ref{fig:motivation},
vivid color and rule of thirds are highly
relevant in rating landscapes but not for closeup portraits.
In~\cite{murray2012ava,xin2015TMM}, contents at the category level are assumed to
be given in both training and testing stages, and category-specific models are
then trained or fine-tuned.
Here we propose to incorporate the category information into our model for joint optimization and prediction,
so that the model can also work on those images with unknown category labels.



We fine-tune the top two layers of AlexNet~\cite{krizhevsky2012imagenet} with
softmax loss to train a content-specific branch to predict category
labels\footnote{Even though category classification uses different features
from those in aesthetics rating, we assume the low-level features can be shared
across aesthetics and category levels.}
(as shown by ContClass layer in Fig.~\ref{fig:net_summary} (c)).
Rather than making a hard category
selection, we use the softmax output as a weighting vector for combining the
scores produced by the category specific branches, each of which is a
concatenation of attribute feature and content-specific features (denoted by
Att$\_$fea and Cont$\_$fea respectively in Fig.~\ref{fig:net_summary} (c)).
This allows for content
categories to be non-exclusive (\eg a photo of an individual in a nature scene
can utilize attributes for either portrait and scenery photos). During training,
When fine-tuning the whole net as in Fig.~\ref{fig:net_summary} (c),
we freeze the content-classification branch and fine-tune the rest network.

\subsection{Implementation Details}
\label{sec:trainingDetail}
We warp images to $256\times 256$ and randomly crop out a $227\times 227$ window to feed into the network.
The initial learning rate is set at 0.0001 for all layers,
and periodically annealed by 0.1.
We set weight decay $1e-5$ and momentum $0.9$.
We use Caffe toolbox~\cite{jia2014caffe} extended with our ranking loss for training all the models.

To train attribute-adaptive layers, we use softmax loss on AVA dataset which only has binary labels for attributes, and
the Euclidean loss on the AADB dataset which has finer-level attribute scores.
We notice that, on the AVA dataset,
our attribute-adaptive branch yields $59.11\%$ AP and $58.73\%$ mAP for
attribute prediction,
which are comparable to the reported results of style-classification model fine-tuned from AlexNet~\cite{xin2015iccv}.
When learning content-adaptive layers on the AVA dataset for classifying eight categories,
we find the content branch yields $59\%$ content classification accuracy on the testing set.
If we fine-tune the whole AlexNet,
we obtain $62\%$ classification accuracy.
Note that we are not pursuing the best classification performance on either attributes or categories.
Rather, our aim is to train reasonable branches that perform well enough to help with image aesthetics rating.

\section{Experimental Results}
\label{sec:exp}

To validate our model for rating image aesthetics, we
first compare against several baselines including the intermediate models
presented in Section~\ref{sec:ourModels},
then analyze the dependence of model
performance on the model parameters and structure, and finally compare
performance of our model with human annotation in rating image
aesthetics.

\subsection{Benchmark Datasets}

\begin{figure}[t]
\centering
   \includegraphics[width=0.75\linewidth]{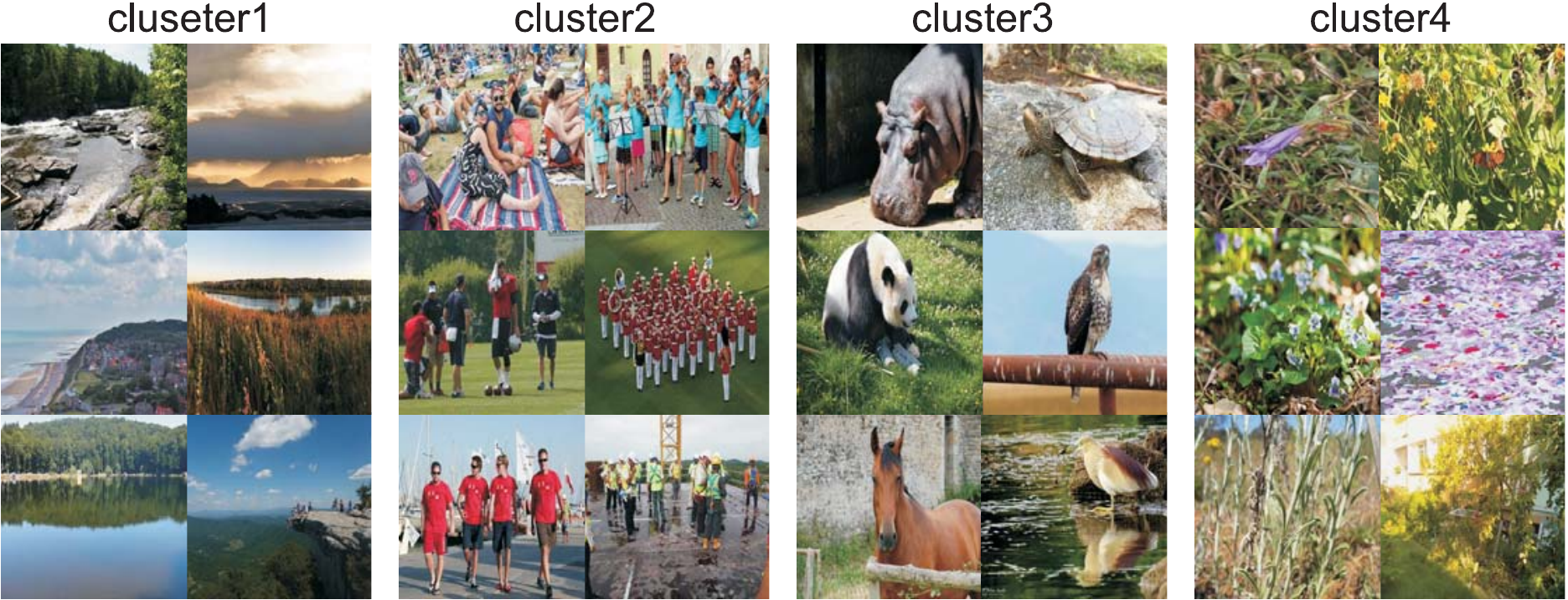}
\vspace{-1mm}
   \caption{Example images from four content clusters found in the training
   set.  These clusters capture thematic categories of image content present in
   AADB without requiring additional manual labeling of training data.}
\label{fig:cluster}
\vspace{-2mm}
\end{figure}

\noindent \textbf{AADB dataset } contains 10,000 images in total, with
detailed aesthetics and attribute ratings, and anonymized raters' identity for specific
images.  We split the dataset into training (8,500), validation (500) and
testing (1,000) sets. Since our dataset does not include ground-truth image content tags, 
we use clustering to find semantic content groups prior to training
content adaptive models.
Specifically, we represent each image using the
\texttt{fc7} features, normalize the feature vector to
be unit Euclidean length, and use unsupervised $k$-means for clustering.  In our
experimental comparison, we cluster training images into $k=10$ content groups,
and transform the distances between a testing image and the centroids into
prediction weights using a softmax.
The value of $k$ was chosen using
validation data (see Section~\ref{ssec:results}).
Fig.~\ref{fig:cluster} shows
samples from four of these clusters,
from which we observe consistencies within each cluster and distinctions across clusters.

\noindent \textbf{AVA dataset } contains approximately 250,000 images,
each of which has about 200 aesthetic ratings ranging on a one-to-ten scale.
For fair comparison,
we follow the experimental practices and train/test split used in
literature~\cite{lu2014rapid,xin2015iccv,murray2012ava} which
results in about 230,000 training and 20,000 test images.  When fine-tuning
AlexNet for binary aesthetics classification, we divide the training set into
two categories (low- and high-aesthetic category), with a score threshold of 5
as used in~\cite{lu2014rapid,xin2015iccv,murray2012ava}.
We use the subset of images which contain style attributes and content tags for training and testing the attribute-adaptive and
content-adaptive branches.

\subsection{Performance Evaluation}
\label{sec:metrics}

To evaluate the aesthetic scores predicted by our model,
we report the ranking correlation
measured by Spearman's $\rho$ between the estimated aesthetics scores and the
ground-truth scores in the test set~\cite{myers2010research}.
Let $r_i$ indicate the rank of the $i$th item when we sort the list by scores
$\{y_i\}$ and ${\hat r}_i$ indicate the rank when ordered by $\{{\hat y}_i\}$.
We can compute the disagreement in the two rankings of a particular element $i$
as $d_i=r_i-{\hat r}_i$.  The Spearman's $\rho$ rank correlation statistic is
calculated as $\rho = 1- \frac{6\sum d_i^2}{ N^3-N}$,
where $N$ is the total number of images ranked.  This correlation coefficient
lies in the range of $[-1,1]$, with larger values corresponding to higher
correlation in the rankings.  The ranking correlation is particularly useful
since it is invariant to monotonic transformations of the aesthetic score
predictions and hence avoids the need to precisely calibrate output scores
against human ratings.
For purposes of comparing to existing classification accuracy results reported
on the AVA dataset, we simply threshold the estimated scores $[\hat y_i > \tau]$
to produce a binary prediction where the threshold $\tau$ is determined on the validation set.



\subsection{Results}
\label{ssec:results}

\begin{table}[t]
\centering
\setlength\tabcolsep{4pt}
\begin{minipage}{0.48\textwidth}
\centering
\small
\centering
\caption{Performance comparison of different models on AADB dataset.}
\begin{tabular}{|l|c|c|c|c|}
\hline
Methods      &			$\rho$ \\
\hline\hline
AlexNet$\_$FT$\_$Conf  & 0.5923 \\
\hline
Reg              & 	0.6239 \\
Reg+Rank (cross-rater)        & 0.6308 \\
Reg+Rank (within-rater)       & 0.6450 \\
Reg+Rank (within- $\&$ cross-)& 0.6515 \\
Reg+Rank+Att    &	 0.6656  \\
Reg+Rank+Cont &	 0.6737  \\
\hline
\hline
\textbf{Reg+Rank+Att+Cont} &	 \textbf{0.6782}  \\
\hline
\end{tabular}
\label{tab:modelPerformanceAADB}
\end{minipage}%
\hfill
\begin{minipage}{0.48\textwidth}
\scriptsize
\centering
\caption{Performance comparison of different models on AVA dataset.}
\begin{tabular}{|l|c|c|c|c|}
\hline
Methods         &			$\rho$&		ACC ($\%$)\\
\hline\hline
Murray \etal~\cite{murray2012ava} &	- & 68.00 \\
SPP~\cite{he2014spatial} & 	- & 72.85 \\
AlexNet$\_$FT$\_$Conf	 & 0.4807 & 71.52\\
\hline
DCNN~\cite{lu2014rapid}  & 	- & 73.25 \\
RDCNN~\cite{lu2014rapid} & 	- & 74.46 \\
RDCNN$\_$semantic~\cite{xin2015TMM}  & 	- & 75.42 \\
DMA~\cite{xin2015iccv}  & 	- & 74.46 \\
DMA$\_$AlexNet$\_$FT~\cite{xin2015iccv} & 	- & 75.41 \\
\hline
Reg              & 	0.4995 & 72.04  \\
Reg+Rank         & 0.5126 & 71.50 \\
Reg+Att          & 0.5331 & 75.32  \\
Reg+Rank+Att     & 0.5445 & 75.48  \\
Reg+Rank+Cont    & 0.5412& 73.37  \\
\hline
\hline
\textbf{Reg+Rank+Att+Cont} &  \textbf{0.5581} & \textbf{77.33}  \\
\hline
\end{tabular}
\label{tab:modelPerformanceAVA}
\end{minipage}
\vspace{-0.2in}
\end{table}

For comparison, we also train a model for binary aesthetics
classification by fine-tuning AlexNet (AlexNet$\_$FT$\_$Conf).  This has
previously been shown to be a strong baseline for aesthetic
classification~\cite{xin2015iccv}.  We use the softmax confidence score
corresponding of the high-aesthetics class as the predicted aesthetic rating.
As described in Section~\ref{sec:ourModels}, we consider variants of
our architecture including the regression network alone (Reg),
along with the
addition of the pairwise ranking loss (Reg+Rank), attribute-constraint
branches (Reg+Rank+Att) and content-adaptive branches (Reg+Rank+Cont).
We also evaluate different pair-sampling strategies including within- and
cross-rater sampling.

\vspace{2mm}
\noindent \textbf{Model Architecture and Loss Functions:}
Table~\ref{tab:modelPerformanceAADB} and~\ref{tab:modelPerformanceAVA}
list the performance on AADB and AVA datasets, respectively.  From these
tables, we notice several interesting observations.  First,
AlexNet$\_$FT$\_$Conf model yields good ranking results measured by $\rho$.
This indicates that the confidence score in softmax can
provide information about relative rankings. 
Second, the regression net outperforms the AlexNet$\_$FT$\_$Conf model, and ranking loss further
improves the ranking performance on both datasets.  This shows the
effectiveness of our ranking loss which considers relative aesthetics
ranking of image pairs in training the model.
More specifically,
we can see from Table~\ref{tab:modelPerformanceAADB} that,
by sampling image pairs according to the the averaged ground-truth scores, \ie cross-rater sampling only,
Reg+Rank (cross-rater) achieves the ranking coefficient $\rho=0.6308$;
whereas if only sampling image pairs within each raters,
we have $\rho=0.6450$ by by Reg+Rank (within-rater).
This demonstrates the effectiveness of sampling image pairs within the same raters,
and validates our idea that the same individual has consistent aesthetics ratings.
When using both strategies to sample image pairs,
the performance is even better by Reg+Rank (within- $\&$ cross-), leading to $\rho=0.6515$.
This is possibly due to richer information contained in more training pairs.
By comparing the results in Table~\ref{tab:modelPerformanceAVA} between ``Reg''
(0.4995) and ``Reg+Rank'' (0.5126), and between ``Reg+Att'' (0.5331) and
``Reg+Rank+Att'' (0.5445) , we clearly observe that the ranking loss improves the ranking correlation.  
In this case, we can only exploit the cross-rater sampling strategy since
rater's identities are not available in AVA for the stronger within-rater
sampling approach.
We note that for values of $\rho$ near $0.5$ computed over $20000$ test images on
AVA dataset, differences in rank correlation of $0.01$ are highly statistically
significant.  These results clearly show that the ranking loss helps enforce
overall ranking consistency.

To show that improved performance is due to the side information (\eg attributes) other than a wider architecture,
we first train an ensemble of eight rating networks (Reg) and average the results, leading to a rho=0.5336 (c.f. Reg+Rank+Att which yields rho=0.5445).
Second, we try directly training the model with a single Euclidean loss using a wider intermediate layer with eight times more parameters. In this case we observed severe overfitting. This suggests for now that the side-supervision is necessary to effectively train such an architecture.

Third, when comparing Reg+Rank with Reg+Rank+Att, and Reg+Rank with Reg+
Rank+Cont, we can see that both attributes and content further improve
ranking performance.  While image content is not annotated on the AADB dataset,
our content-adaptive model based on unsupervised $K$-means clustering still
outperforms the model trained without content information.  The performance
benefit of adding attributes is substantially larger for AVA than AADB.
We expect this is due to (1) differences in the definitions of attributes between
the two datasets, and (2) the within-rater sampling for AADB, which already
provides a significant boost making further improvement using attributes more
difficult.  The model trained with ranking loss, attribute-constraint and
content-adaptive branches naturally performs the best among all models.  It is
worth noting that, although we focus on aesthetics ranking during training, we
also achieve the state-of-the-art binary classification accuracy in AVA.  This
further validates our emphasis on relative ranking, showing that learning to
rank photo aesthetics can naturally lead to good classification performance.



\begin{table}[t]
\centering
\setlength\tabcolsep{4pt}
\begin{minipage}{0.48\textwidth}
\centering
\small
\centering
\caption{Ranking performance $\rho$ vs. rank loss weighting $\omega_r$ in Eq.~\ref{eq:regRankModel}.}
\begin{tabular}{|c|c|c|c|c|}
\hline
$\omega_r$& $0.0$ & 0.1 & $1$ & $2$ \\
\hline\hline
AADB    & $0.6382$ & $0.6442$ & $0.6515$ & $0.6276$ \\
AVA     & $0.4995$ & 0.5126 & $0.4988$ & $0.4672$ \\
\hline
\end{tabular}
\label{tab:perf_vs_ranklossWeight}
\end{minipage}%
\hfill
\begin{minipage}{0.48\textwidth}
\small
\centering
\caption{Ranking performance ($\rho$) of ``Reg+Rank'' with different numbers of sampled image pairs on AADB dataset.}
\begin{tabular}{|c|c|c|c|c|}
\hline
	$\#$ImgPairs      &  2 million & 5 million \\
\hline
cross-rater & 0.6346 &  0.6286 \\
within-rater & 0.6450 &  0.6448 \\
within- $\&$ cross-rater & 0.6487 & 0.6515 \\
\hline
\end{tabular}
\label{tab:Perf_vs_numSampledPairs}
\end{minipage}
\vspace{-0.2in}
\end{table}

\begin{figure}[t]
  \centering
  \begin{minipage}[b]{0.43\textwidth}
    \includegraphics[width=0.990\linewidth]{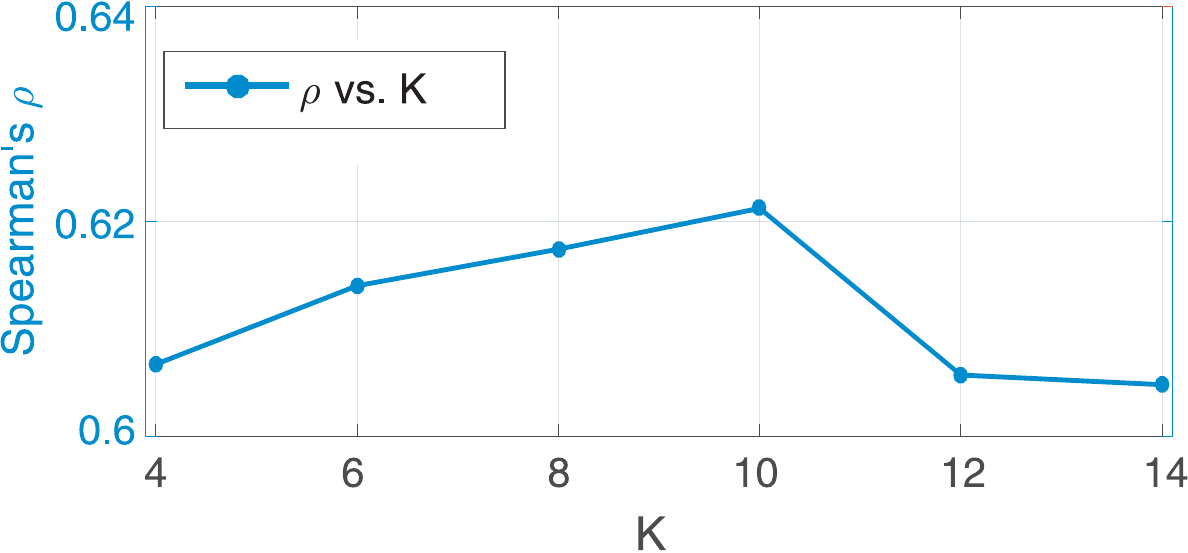}
    \caption{Dependence of model performance by varying the number of content
   clusters.  We select $K=10$ clusters in our experiments on AADB.}
\label{fig:performance_vs_K}
  \end{minipage}
  \hfill
  \begin{minipage}[b]{0.55\textwidth}
    \includegraphics[width=0.99\linewidth]{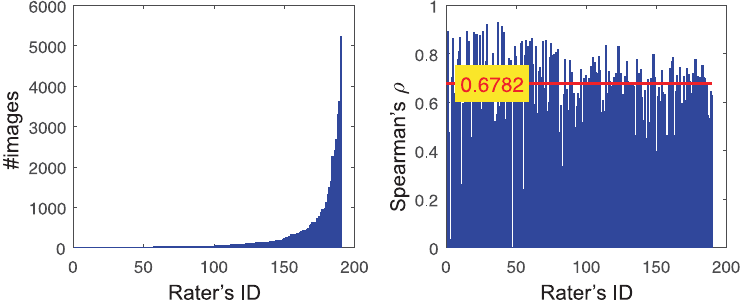}
    \caption{Panels show (left) the number of images labeled by each worker,
   and the performance of each individual rater w.r.t Spearman's $\rho$ (Right).
   Red line shows our model's performance.}
\label{fig:workerPerformanceDistribution}
  \end{minipage}
\end{figure}

\vspace{2mm}
\noindent \textbf{Model Hyperparameters:}
In training our content-adaptive model on the AADB dataset which lacks supervised
content labels, the choice of cluster number is an important parameter.
Fig.~\ref{fig:performance_vs_K} plots the $\rho$ on validation data
as a function of the number of clusters $K$ for the Reg+Cont model (without
ranking loss).  We can see the finer clustering improves performance as each
content specific model can adapt to a sub-category of images.
However, because
the total dataset is fixed, performance eventually drops as the amount of
training data available for tuning each individual content-adaptive branch
decreases. We thus fixed $K=10$ for training our unified network on AADB.

The relative weightings of the loss terms (specified by $\omega_r$ in Eq.~\ref{eq:regRankModel})
is another important parameter.  Table~\ref{tab:perf_vs_ranklossWeight} shows the ranking
correlation test performance on both datasets w.r.t. different choices of
$\omega_r$.
We observe that larger $\omega_r$ is favored in AADB than that in AVA,
possibly due to the contribution from the within-rater image pair sampling strategy.
We set $\omega_a$ (in Eq.~\ref{eq:regRankAttModel})
to $0.1$ for jointly fine-tuning attribute regression and aesthetic rating.
For the rank loss, we used validation performance to set
the margin $\alpha$ to $0.15$ and $0.02$ on AVA and AADB respectively.

\vspace{2mm}
\noindent \textbf{Number of Sampled Image Pairs:}
Is it possible that better performance can be obtained through more sampled pairs instead of leveraging rater's information?
To test this,
we sample 2 and 5 million image pairs given the fixed training images on the AADB dataset,
and report in Table~\ref{tab:Perf_vs_numSampledPairs} the performance of model ``Reg+Rank'' using different sampling strategies,
\ie within-rater only, cross-rater only and within-$\&$cross-rater sampling.
It should be noted the training image set remains the same, we just sample more pairs from them.
We can see that adding more training pairs yields little differences in the final results,
and even declines slightly when using higher cross-rater sampling rates.
These results clearly emphasize the effectiveness of our proposed sampling
strategy which (perhaps surprisingly) yields much bigger gains than simply
increasing the number of training pairs by 2.5x.

\vspace{2mm}
\noindent \textbf{Classification Benchmark Performance:}
Our model achieves state-of-the-art classification performance on the AVA
dataset simply by thresholding the estimated score (Table~\ref{tab:modelPerformanceAVA}).
It is worth noting that our model uses only the whole warped down-sampled images
for both training and testing, without using any high-resolution patches from
original images.  Considering the fact that the fine-grained information
conveyed by high-resolution image patches is especially useful for image
quality assessment and aesthetics
analysis~\cite{lu2014rapid,kang2014convolutional,xin2015iccv}, it is quite
promising to see our model performing so well.  The best reported
results~\cite{xin2015iccv} for models that use low resolution warped images for
aesthetics classification are based on
Spatial Pyramid Pooling Networks (SPP)~\cite{he2014spatial}
and achieves an
accuracy of $72.85\%$.  Compared to SPP,
our model achieves $77.33\%$, a gain of $4.48\%$, even though our model is not
tuned for classification.
Previous work~\cite{kang2014convolutional,lu2014rapid,xin2015iccv} has shown
that leveraging the high-resolution patches could lead to additional $5\%$
potential accuracy improvement.
We expect a further accuracy boost would be possible by applying this strategy with our model.


\begin{table}[t]
\centering
\setlength\tabcolsep{4pt}
\begin{minipage}{0.48\textwidth}
\centering
\small
\centering
\caption{Human perf. on the AADB dataset.}
\begin{tabular}{|l|l|c|c|c|c|}
\hline
$\#$images & $\#$workers  & $\rho$ \\
\hline\hline
$>$0 &  190  &  0.6738 \\
\hline
$>$100 &  65 &  0.7013 \\
\hline
$>$200 &  42 &  0.7112 \\
\hline\hline
Our best & -- &  0.6782 \\
\hline
\end{tabular}
\label{tab:HumanWorkerComparison}
\end{minipage}%
\hfill
\begin{minipage}{0.48\textwidth}
\small
\centering
\caption{Cross dataset train/test evaluation.}
\begin{tabular}{p{0.03cm}lcc}
\hline
\multicolumn{2}{c}{\multirow{2}{*}{Spearman's $\rho$}}            & \multicolumn{2}{c} {test} \\
\cmidrule(r){3-4}
&            &	AADB&		AVA \\
\hline
\parbox[t]{2mm}{\multirow{2}{*}{\rotatebox[origin=c]{90}{train}}} &   AADB        &	0.6782 &	0.1566	\\
                                                                    &   AVA         &	0.3191 &	0.5154	\\
\hline
\end{tabular}
\label{tab:crossDBtest}
\end{minipage}
\end{table}

\subsection{Further Comparison with Human Rating Consistency}

We have shown that our model achieves a high level of agreement with average
aesthetic ratings and outperforms many existing models.  The raters' identities
and ratings for the images in our AADB dataset enable us to further analyze
agreement between our model each individual as well as intra-rater consistency.
While human raters produce rankings which are similar with high statistical
significance, as evaluated in Section~\ref{ssec:datasetCollection}, there is
variance in the numerical ratings between them.

To this end, we calculate ranking correlation $\rho$
between each individual's ratings and the ground-truth average score.
When comparing an individual to the ground-truth, we do not exclude that
individual's rating from the ground-truth average for the sake of comparable
evaluations across all raters.
Fig.~\ref{fig:workerPerformanceDistribution} shows
the number of images each rater has rated and their corresponding performance
with respect to other raters.  Interestingly, we find that the
hard workers tend to provide more consistent ratings.
In Table~\ref{tab:HumanWorkerComparison}, we summarize the individuals'
performance by choosing a subset raters based on the number of images they have
rated.  This clearly indicates that the different human raters annotate the images
consistently, and when labeling more images, raters contribute more
stable rankings of the aesthetic scores.

Interestingly, from Table~\ref{tab:HumanWorkerComparison},
we can see that our
model actually performs above the level of human consistency (as measured by
$\rho$) averaged across all workers.  However, when concentrating on the
``power raters'' who annotate more images, we still see a gap between machine and
human level performance in terms of rank correlation $\rho$.



\subsection{Cross-Dataset Evaluation}
\label{ssec:transferability}

As discussed in Section~\ref{ssec:datasetCollection},
AVA contains
professional images downloaded from a community based rating website; while our
AADB contains a much more balanced distribution of consumer photos and professional photos rated by AMT workers,
so has better generalizability to wide range of real-world photos.

To quantify the differences between these
datasets, we evaluate whether models trained on one dataset perform well
on the other.  Table~\ref{tab:crossDBtest} provides a comparison of the
cross-dataset performance.  Interestingly, we find the models trained on either
dataset have very limited ``transferability''.
We conjecture there are two reasons.  First, different groups of raters have
different aesthetics tastes.  This can be verified that, when looking at the
DPChallenge website where images and ratings in the AVA dataset were taken
from.  DPChallenge provides a breakdown of scores which shows notable
differences between the average scores among commenters, participants and
non-participants.  Second, the two datasets contain photos with different
distributions of visual characteristics.  For example, many AVA photos are
professionally photographed or heavily edited; while AADB contains many daily
photos from casual users.  This observation motivates the need for further
exploration into mechanisms for learning aesthetic scoring that is adapted to
the tastes of specific user groups or photo
collections~\cite{caicedo2011collaborative}.

\section{Conclusion}
We have proposed a CNN-based method that unifies photo style attributes and
content information to rate image aesthetics.  In training this architecture,
we leverage individual aesthetic rankings which are provided by a novel dataset
that includes aesthetic and attribute scores of multiple images by individual
users. We have shown that our model is also effective on existing
classification benchmarks for aesthetic judgement. Despite not using
high-resolution image patches, the model achieves state-of-the-art
classification performance on the AVA benchmark by simple thresholding.
Comparison to individual raters suggests that our model performs as well as the
``average'' mechanical turk worker but still lags behind more consistent
workers who label large batches of images.
These observations suggest future work in developing aesthetic rating systems that can adapt to individual user preferences.

\bibliographystyle{splncs03}
\bibliography{egbib}

\clearpage

\section*{Appendix: Aesthetics and Attributes Database (AADB)}
\label{sec:AADB}

\subsection*{Attributes in AADB}
We select eleven attributes that are highly related to image aesthetics after consulting professional photographers,
which are
\begin{enumerate}
  \item ``balancing element'' -- whether the image contains balanced elements;
  \item ``content'' -- whether the image has good/interesting content;
  \item ``color harmony'' -- whether the overall color of the image is harmonious;
  \item ``depth of field'' -- whether the image has shallow depth of field;
  \item ``lighting'' -- whether the image has good/interesting lighting;
  \item ``motion blur'' -- whether the image has motion blur;
  \item ``object emphasis'' -- whether the image emphasizes foreground objects;
  \item ``rule of thirds'' -- whether the photography follows rule of thirds;
  \item ``vivid color'' -- whether the photo has vivid color, not necessarily harmonious color;
  \item ``repetition'' -- whether the image has repetitive patterns;
  \item ``symmetry'' -- whether the photo has symmetric patterns.
\end{enumerate}
These attributes span traditional photographic principals of color, lighting,
focus and composition, and provide a natural vocabulary for use in applications,
such as auto photo editing and image retrieval.
To visualize images containing these attributes,
please refer to the attached our AMT instruction in the end of this supplementary material.
The instruction is used for teaching raters to pass the qualification test.

\subsection*{Data Collection By Amazon Mechanical Turk}
\begin{figure}
\begin{center}
   \includegraphics[width=0.5\linewidth]{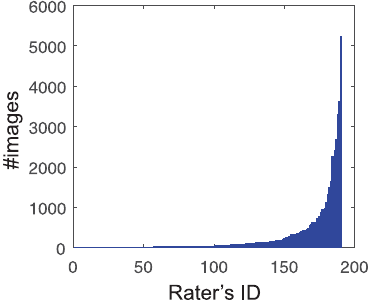}
\end{center}
   \caption{Long tail distribution of AMT workers: number of rated images vs. each worker.}
\label{fig:longTailDis}
\end{figure}

To collect a varied set of photographic images, we download images
from Flickr website\footnote{www.flickr.com},
which carry a Creative Commons
license.
We manually curate the dataset to remove non-photographic images
(\eg cartoons, drawings, paintings, ads images, adult-content images, etc.).
We have multiple workers independently annotate each image with an overall
aesthetic score and the eleven meaningful attributes using Amazon
Mechanical Turk\footnote{www.mturk.com}.

\begin{figure*}[!ht]
\begin{center}
   \includegraphics[width=0.990\linewidth]{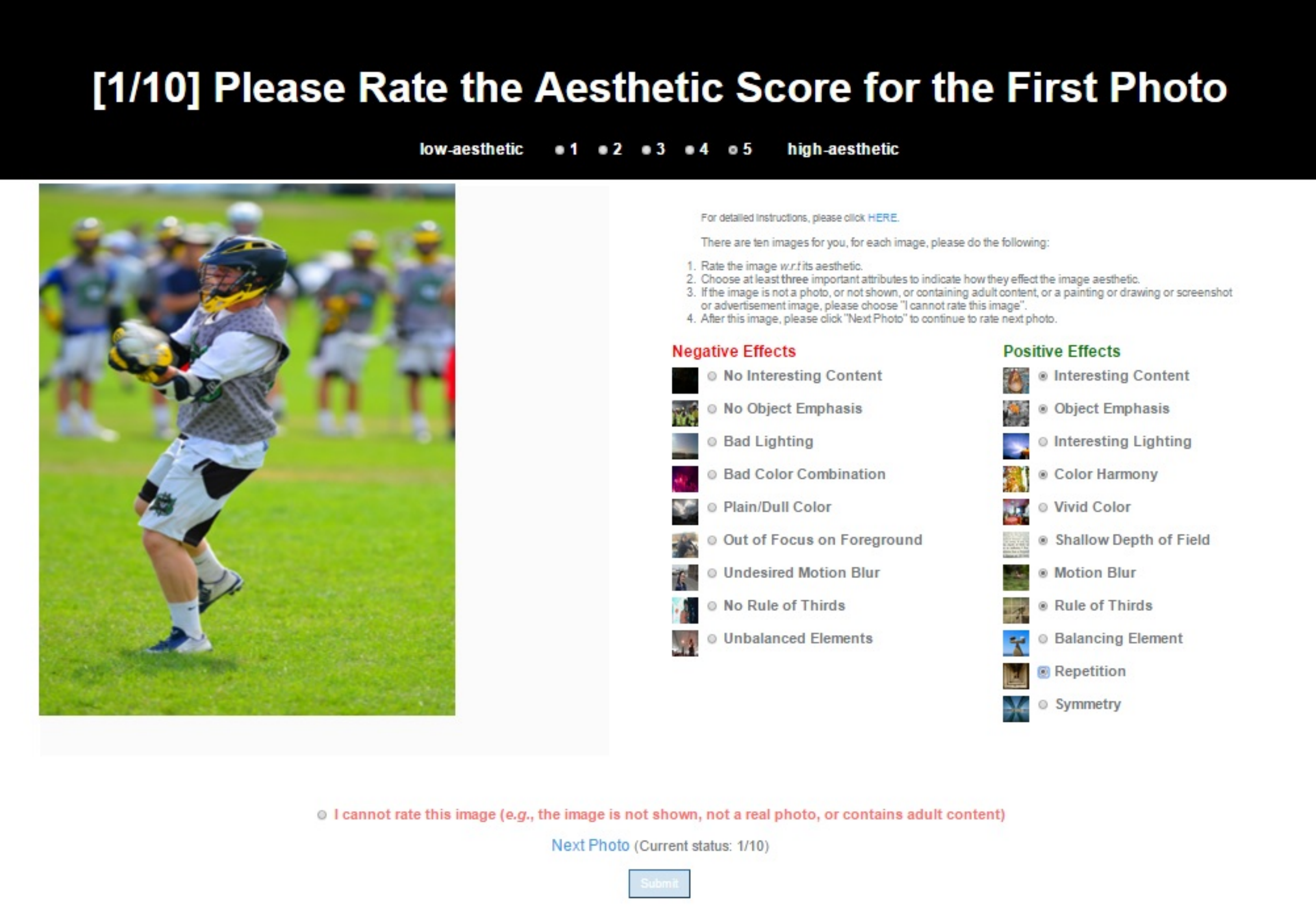}
\end{center}
   \caption{Interface of data collection by AMT.}
\label{fig:demo_AMT}
\end{figure*}

For each attribute,
we allow workers to click ``positive'' if this attribute conveyed by the image can enhance the image aesthetic quality, or ``negative'' if the attribute degrades image aesthetics.
The default is ``null'', meaning the attribute does not effect image aesthetics.
For example,
``positive'' vivid color means the vividness of the color presented in the image has a positive effect on the image aesthetics;
while the counterpart ``negative'' means, for example, there is dull color composition.
Note that we do not let workers tag negative repetition and symmetry,
as for the two attributes negative values do not make sense.

We launch a task consisting of 10,000 images on AMT, and let five different workers label each image.
All the workers must read instructions and pass a qualification exam before they become qualified to do our task.
The images are split into batches,
each of which contains ten images.
Therefore,
raters will annotate different numbers of batches.
There are 190 workers in total doing our AMT task,
and the workers follow long tail distribution,
as demonstrated by Figure~\ref{fig:longTailDis}.
Figure~\ref{fig:demo_AMT} shows the interface of our AMT task.

\begin{figure}[!ht]
\begin{center}
   \includegraphics[width=0.60\linewidth]{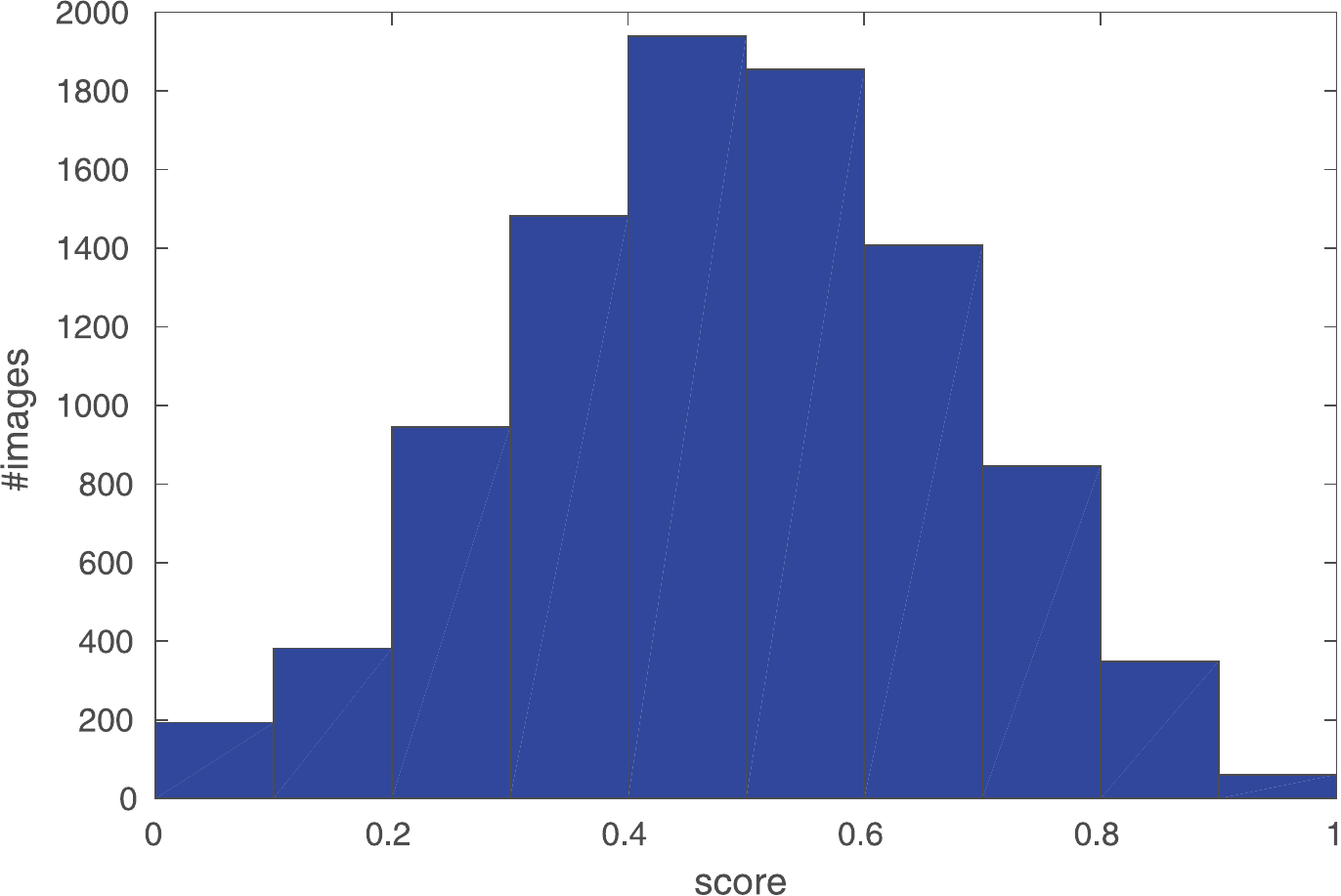}
\end{center}
   \caption{The distribution of rated image aesthetic scores by the AMT workers follows a Gaussian distribution.}
\label{fig:overallScoreDistribution}
\end{figure}

Note that even though judging these attributes is also subjective,
the averaged scores of these attributes indeed reflect good information if we visualize the ranked images w.r.t averaged scores.
Therefore,
we use the averaged score as the ground truth, for both aesthetic score and attributes.
Furthermore,
we normalize aesthetic score to the range of $[0,1]$, as shown by Figure~\ref{fig:overallScoreDistribution},
from which we can see  that ratings
are well fit by a Gaussian distribution.
This observation is consistent with that reported in \cite{murray2012ava}.
In our experiments
we  normalize the attributes' scores to the range of $[-1,1]$.
The images are split into testing set (1,000 images), validation set (500 images) and training set (the rest).

\section*{Appendix: Statistics of AADB}
\label{sec:statAADB}

\begin{figure*}[!th]
\begin{center}
   \includegraphics[width=0.99\linewidth]{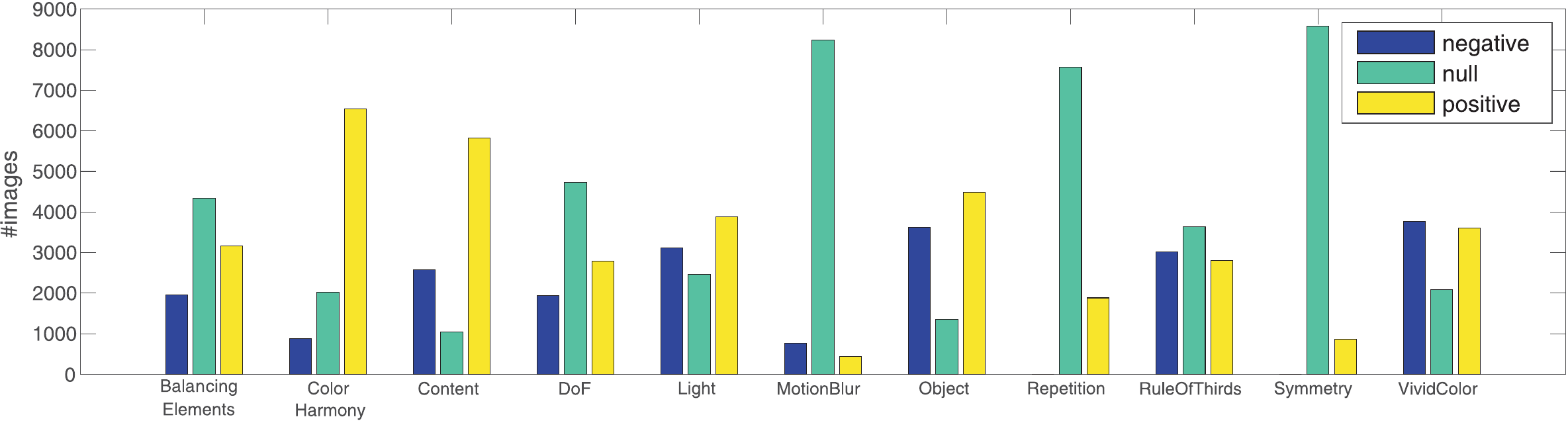}
\end{center}
   \caption{The distributions of all the eleven attributes.
   Note that for attributes repetition and symmetry,
   we do not let AMT workers annotate negative labels, as these attributes are of neutral meaning.
   Instead, we only allow them to point out whether there exist repetition or symmetry.
   To solve the data imbalance problem in training attribute classifiers, we adopt some data augmentation tricks to sample more rare cases.}
\label{fig:AADB_distribution}
\end{figure*}

The final AADB dataset
contains  10,000 images in total, each of which has aesthetic quality
ratings and attribute assignments provided by five different individual raters.
Therefore, we have rating scores for attributes as well,
which is different from AVA dataset~\cite{murray2012ava} in which images only have binary labels for the attributes.
Figure~\ref{fig:AADB_distribution} shows the distribution of each attributes.

\section*{Appendix: Consistency Analysis}
\label{sec:consistencyAnalysis}

\begin{figure}[t]
\centering
   \includegraphics[width=0.9\linewidth]{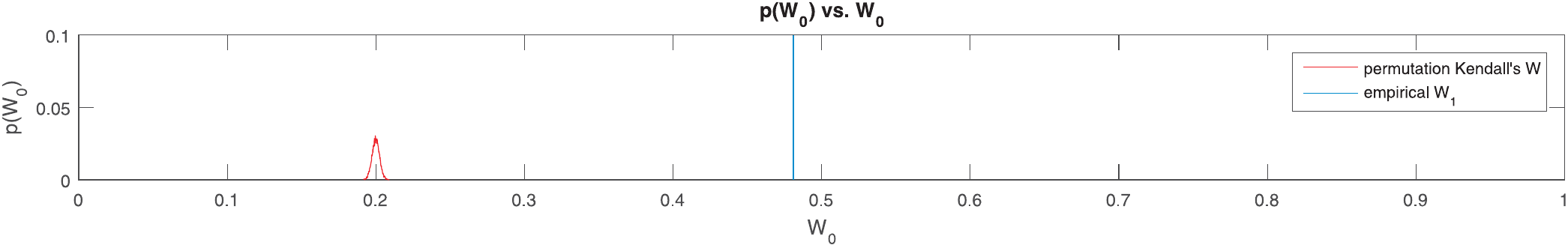}
\vspace*{-4mm}
   \caption{Permutation test on Kendall's $W$: $p(W)$ vs. $W$.}
\label{fig:pW0_vs_W0}
\end{figure}

\begin{figure}[t]
\centering
   \includegraphics[width=0.9\linewidth]{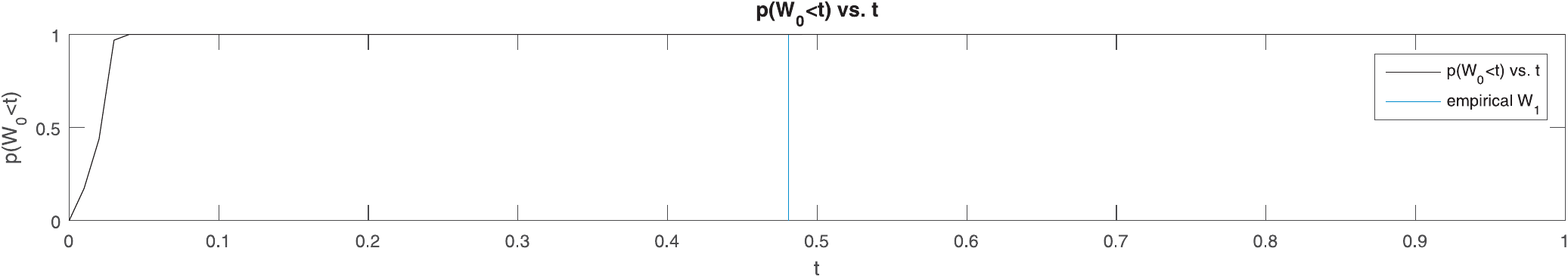}
\vspace*{-4mm}
   \caption{Permutation test on Kendall's $W$: $p(W<t)$ vs. $t$.}
\label{fig:pW0t_vs_t}
\end{figure}

As there are five individuals rating each image,
one may argue that the annotations are not reliable for this subjective task.
Therefore,
we carry out consistency analysis.
We use both Kendall's $W$ and Spearman's $\rho$ for the analysis.
Kendall's $W$ directly measures the agreement among multiple raters, and accounts for tied ranks.
It ranges from 0 (no agreement) to 1 (complete agreement).
Spearman's $\rho$ is used in our paper that compares a pair of ranking lists.

First,
we conduct a permutation test over global $W$ to obtain the distribution of $W$ under the null hypothesis.
We plot the curve of $W$: $p(W)$ vs. $W$ in Fig.~\ref{fig:pW0_vs_W0} and  $p(W<t)$ vs. $t$ in Fig~\ref{fig:pW0t_vs_t}.
We can easily see that the empirical Kendall's $W$ on our AADB dataest is statistically significant.

Then,
for each batch,
we can also evaluate the annotation consistency with Kendall's $W$,
which directly calculates the agreement among multiple raters,
and accounts for tied ranks.
As there are ten images and only five possible ratings for each image,
tied ranks may happen in a batch.
The average Kendall's $W$ over all batches is $0.5322$.
This shows significant consistency of the batches annotated by the AMT workers.
To test the statistical significance of Kendall's $W$ at batch level,
we adopt the Benjamini-Hochberg procedure to control the false discovery rate (FDR) for multiple comparisons~\cite{benjamini2001control}.
At level $Q=0.05$,
$99.07\%$ batches from $1,013$ in total have significant agreement.
This shows that almost all the batches annotated by AMT workers have consistent labels and are reliable for scientific use.

Furthermore,
we can also test the statistical significance w.r.t Spearman's $\rho$ at batch levels using Benjamini-Hochberg procedure.
The $p$-values of pairwise ranks of raters in a batch can be computed by the exact permutation distributions.
We average the pairwise $p$-values as the $p$-value for the batch.
With the FDR level $Q=0.05$,
we find that $98.45\%$ batches have significant agreement.
This further demonstrates the reliability of the annotations.

\section*{Appendix: Analysis of Content-Aware Model}
\label{sec:contAwareModelAnalysis}
\begin{table}
\caption{Analysis of content-aware model on AVA dataset.}
  \centering
  \begin{tabular}{ | c| c | c| c| c| c| c| c| c| c| }
     \hline
         method        & concatGT & concatPred & avg. & weightedSum & weightedSum$\_$FT \\
     \hline\hline
     Spearman's $\rho$ & 0.5367  &  0.5327 & 0.5336  & 0.5335 & 0.5426  \\
     accuracy($\%$)    & 75.41   & 75.33   & 75.39   & 75.33  & 75.57   \\
     \hline
   \end{tabular}
\label{tab:contAwareModelAnalysis}
\end{table}
To show the effectiveness of utilizing content information as a weights for output scores by different content-specific aesthetics rating branches,
we report the performance on AVA dataset of different methods in  Table~\ref{tab:contAwareModelAnalysis}.
Our first method is named ``concatGT'',
which means we use the ground-truth content label of an image,
and get the estimated aesthetic score by the content-specific branch;
then we put all the estimated scores together to get the global Spearman's $\rho$ and classification accuracy.
In method ``concatPred'',
we use the predicted content label to choose which category-specific branch to use for estimating aesthetic score,
then use the same procedure as in ``concatGT''.
In method ``avg.'',
we use all the content-specific aesthetics rating branches to get multiple scores,
and average them to a single score as the final estimation.
In ``weightedSum'',
we use the classification confidence score output by softmax of the content classification branch to do weighted sum for the final score.
In ``weightedSum$\_$FT'',
we fine-tune the whole network but freezing the classification branch,
and use the fine-tuned model to do weighted sum on the scores for the final aesthetics rating.
From this table,
we can clearly observe that ``weightedSum$\_$FT'' performs the best,
which is the one described in the paper.

\section*{Appendix: Demonstration of Our Model}
\label{sec:visualDemo}
\begin{figure*}
    \centering
    \begin{minipage}{.5\textwidth}
        \centering
        \includegraphics[width=0.99\linewidth]{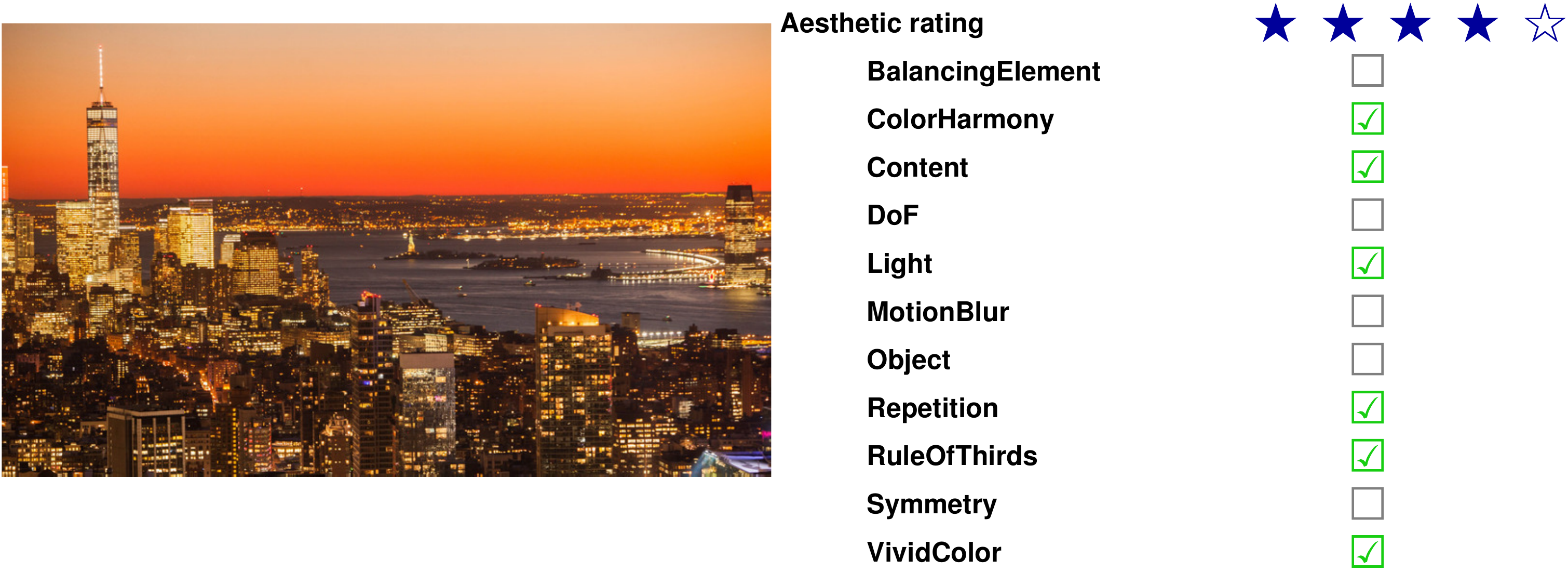}
    \end{minipage}%
    \begin{minipage}{0.5\textwidth}
        \centering
        \includegraphics[width=0.99\linewidth]{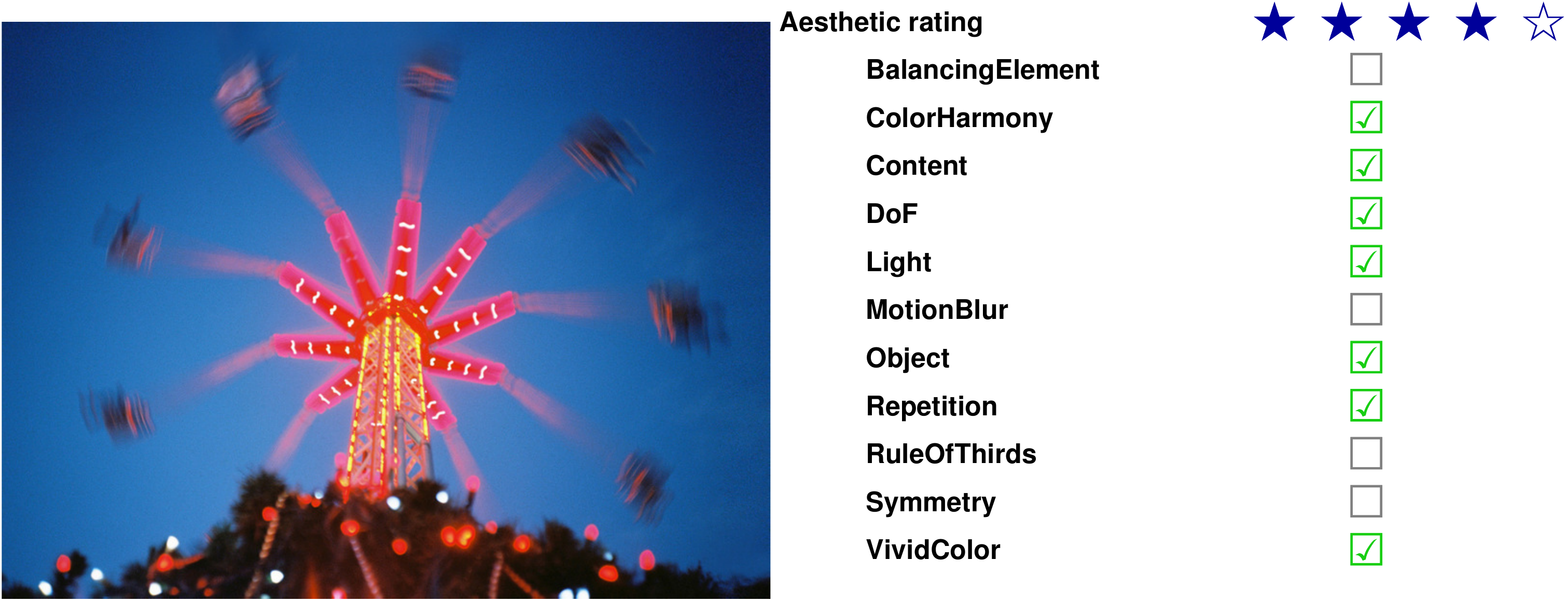}
    \end{minipage}

    \begin{minipage}{0.5\textwidth}
        \centering
        \includegraphics[width=0.99\linewidth]{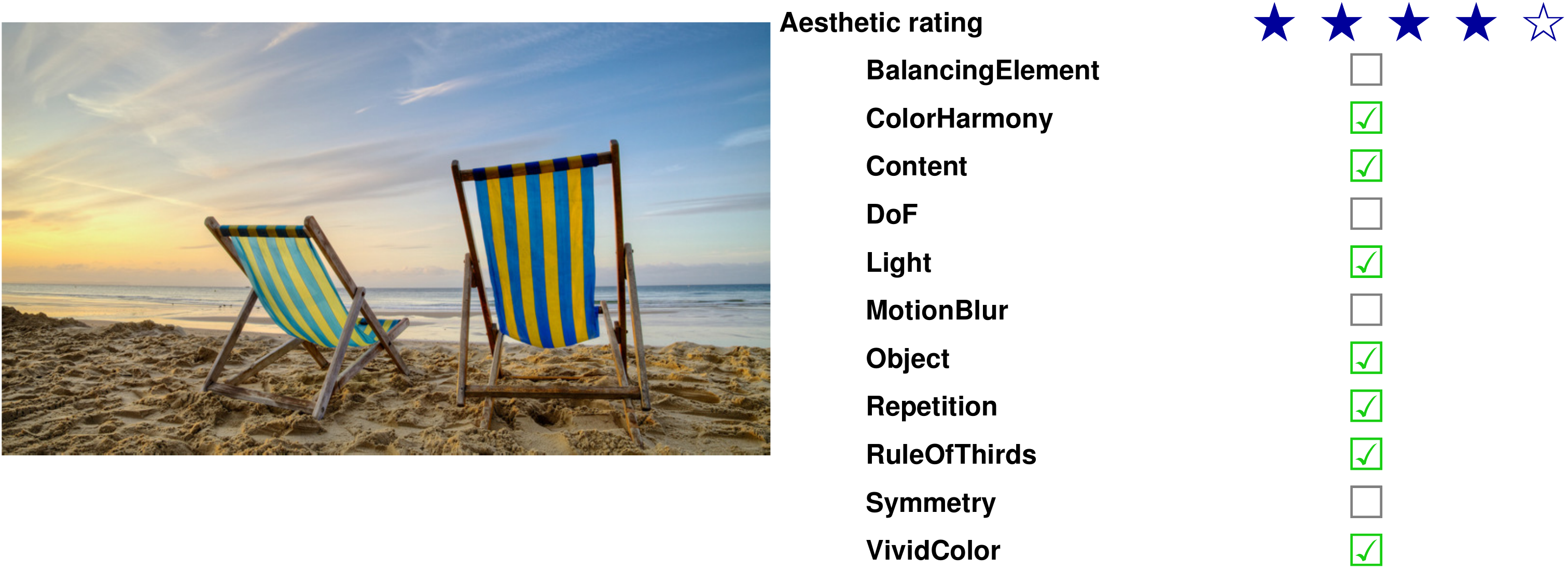}
    \end{minipage}%
    \begin{minipage}{0.5\textwidth}
        \centering
        \includegraphics[width=0.99\linewidth]{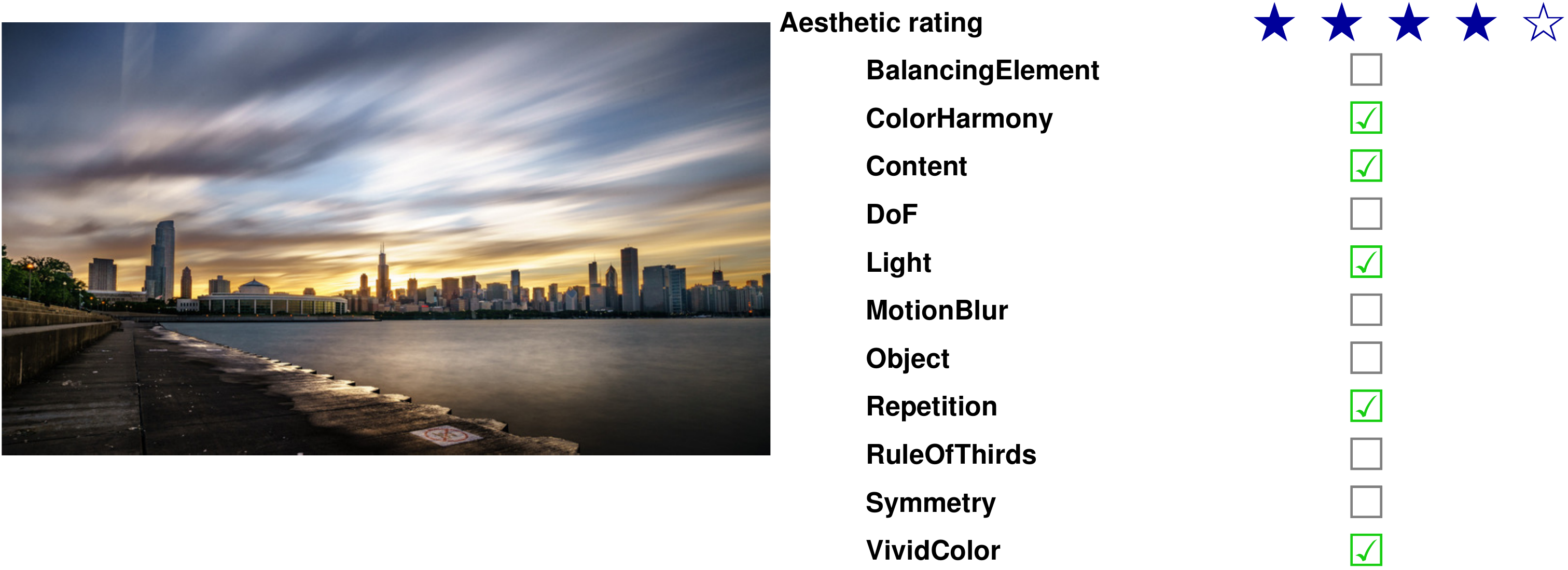}
    \end{minipage}

    \begin{minipage}{0.5\textwidth}
        \centering
        \includegraphics[width=0.99\linewidth]{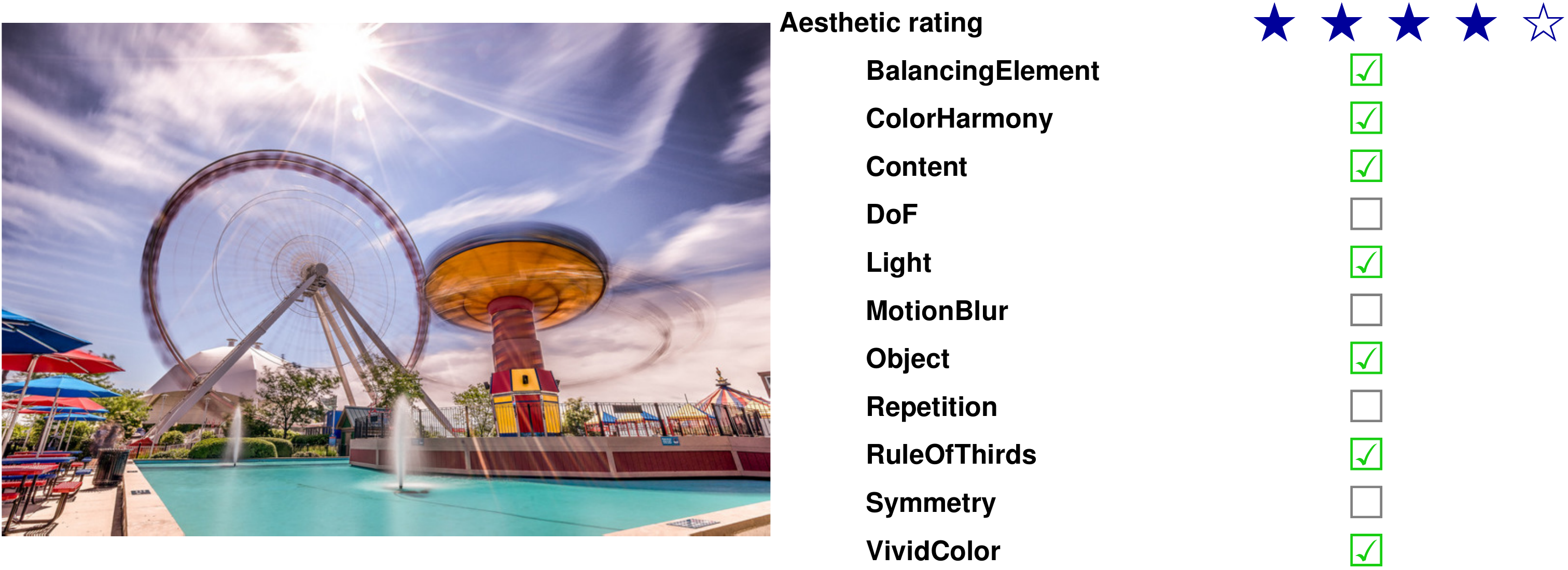}
    \end{minipage}%
    \begin{minipage}{0.5\textwidth}
        \centering
        \includegraphics[width=0.99\linewidth]{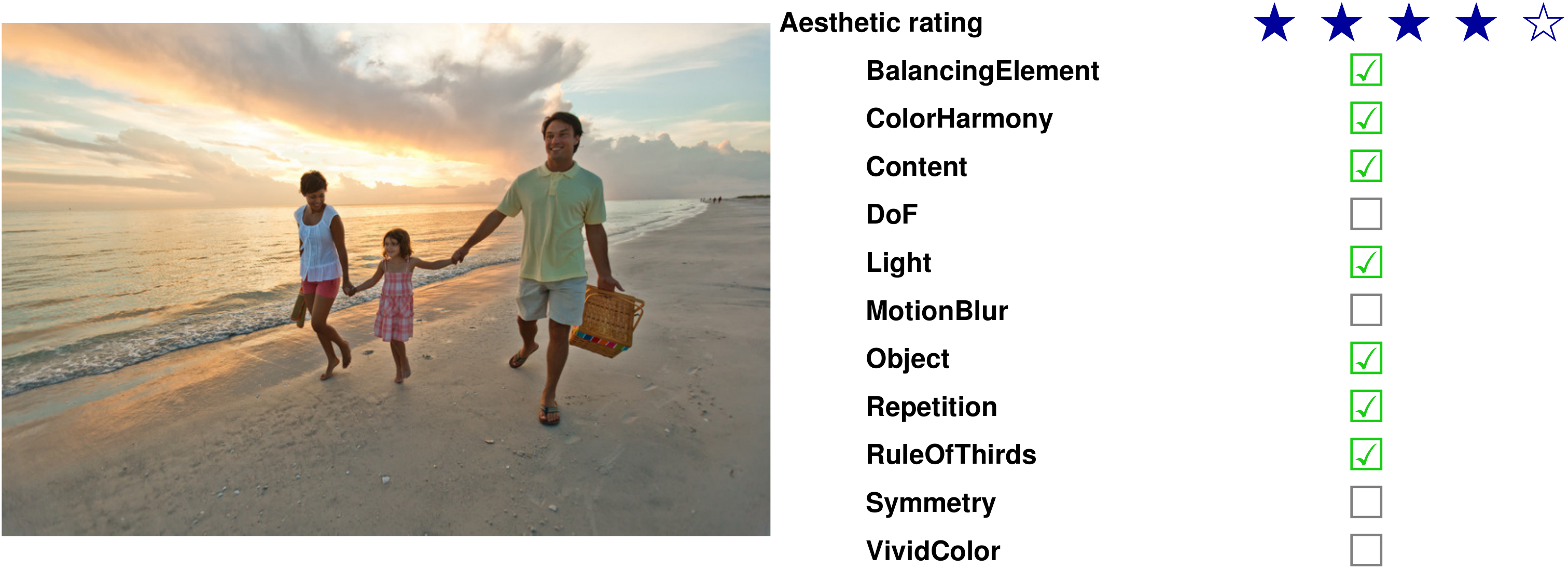}
    \end{minipage}

    \begin{minipage}{0.5\textwidth}
        \centering
        \includegraphics[width=0.99\linewidth]{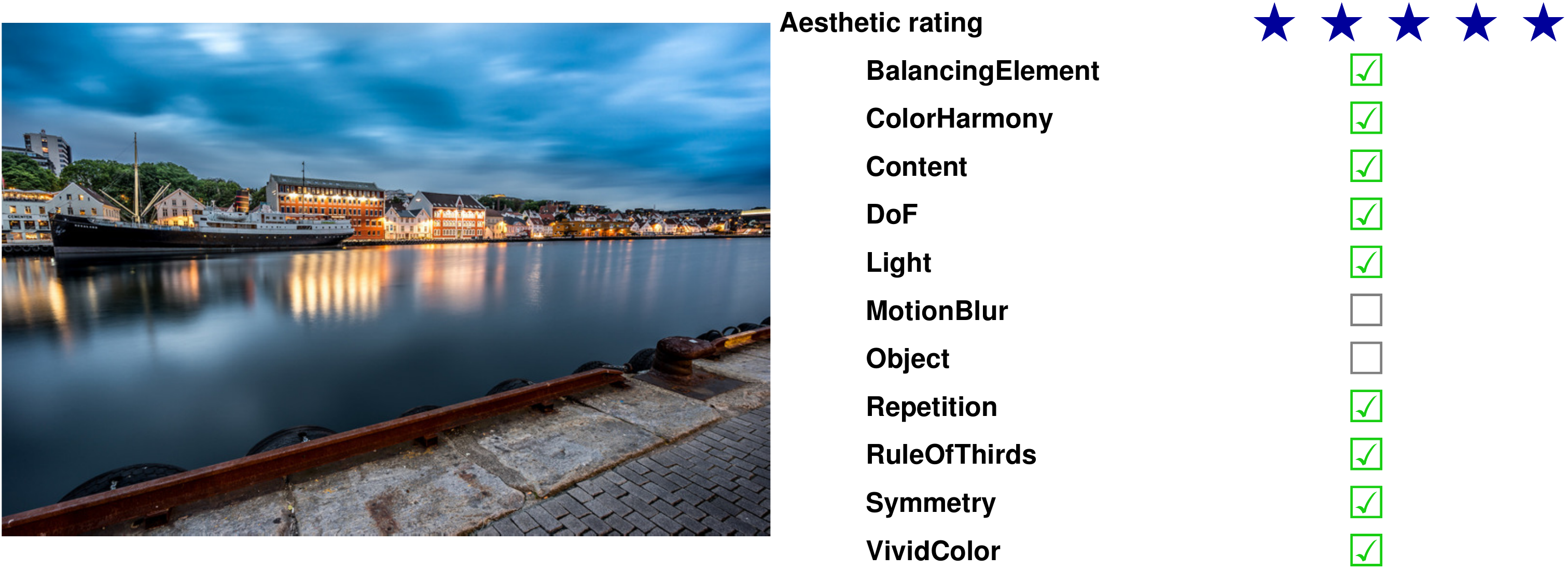}
    \end{minipage}%
    \begin{minipage}{0.5\textwidth}
        \centering
        \includegraphics[width=0.99\linewidth]{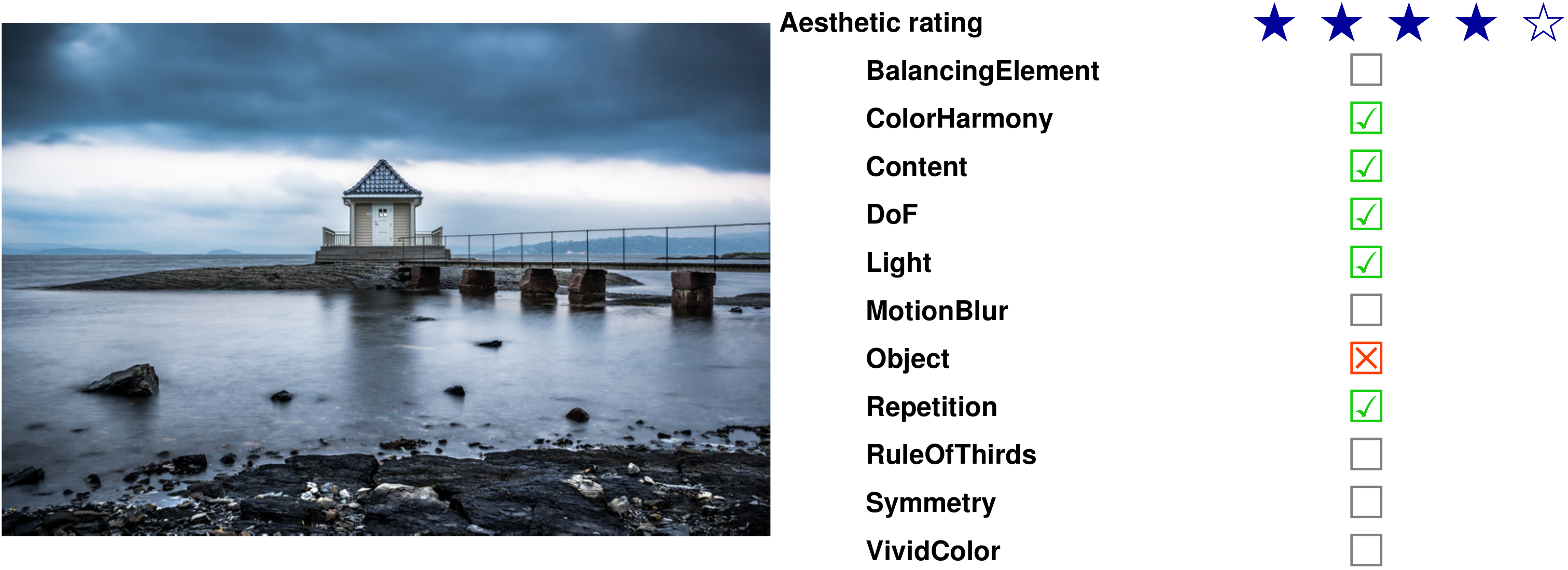}
    \end{minipage}

    \begin{minipage}{0.5\textwidth}
        \centering
        \includegraphics[width=0.99\linewidth]{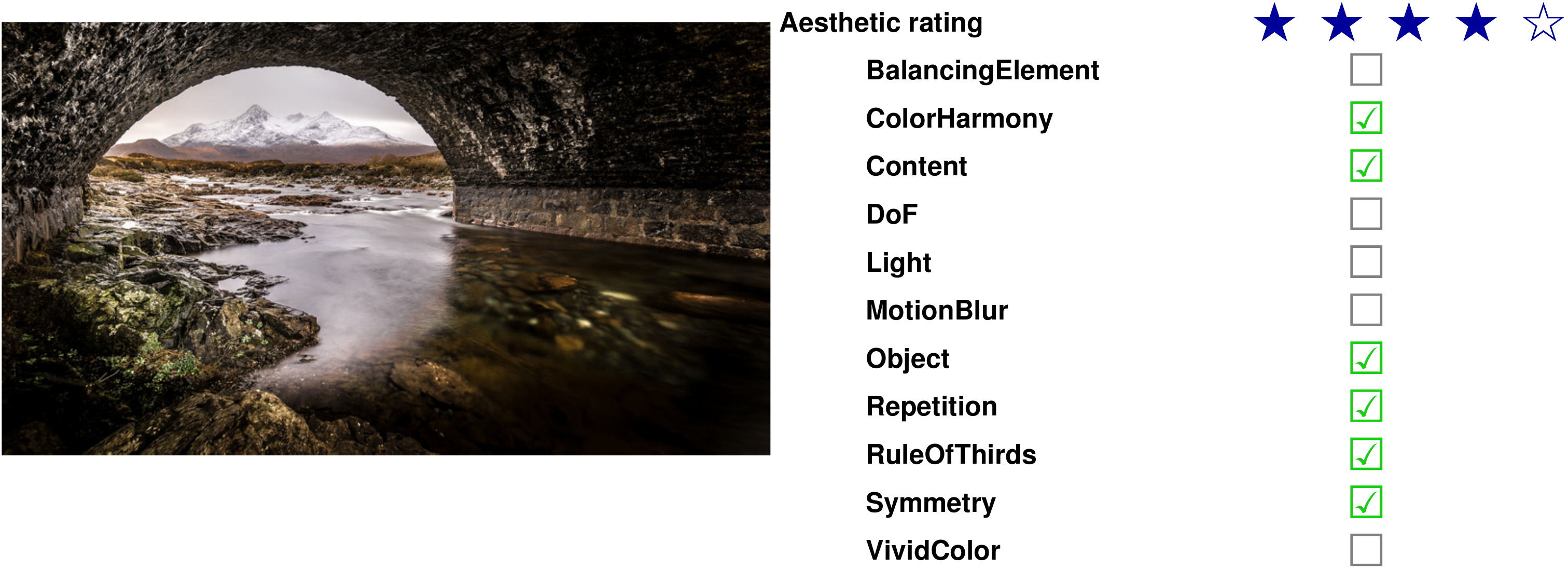}
    \end{minipage}%
    \begin{minipage}{0.5\textwidth}
        \centering
        \includegraphics[width=0.99\linewidth]{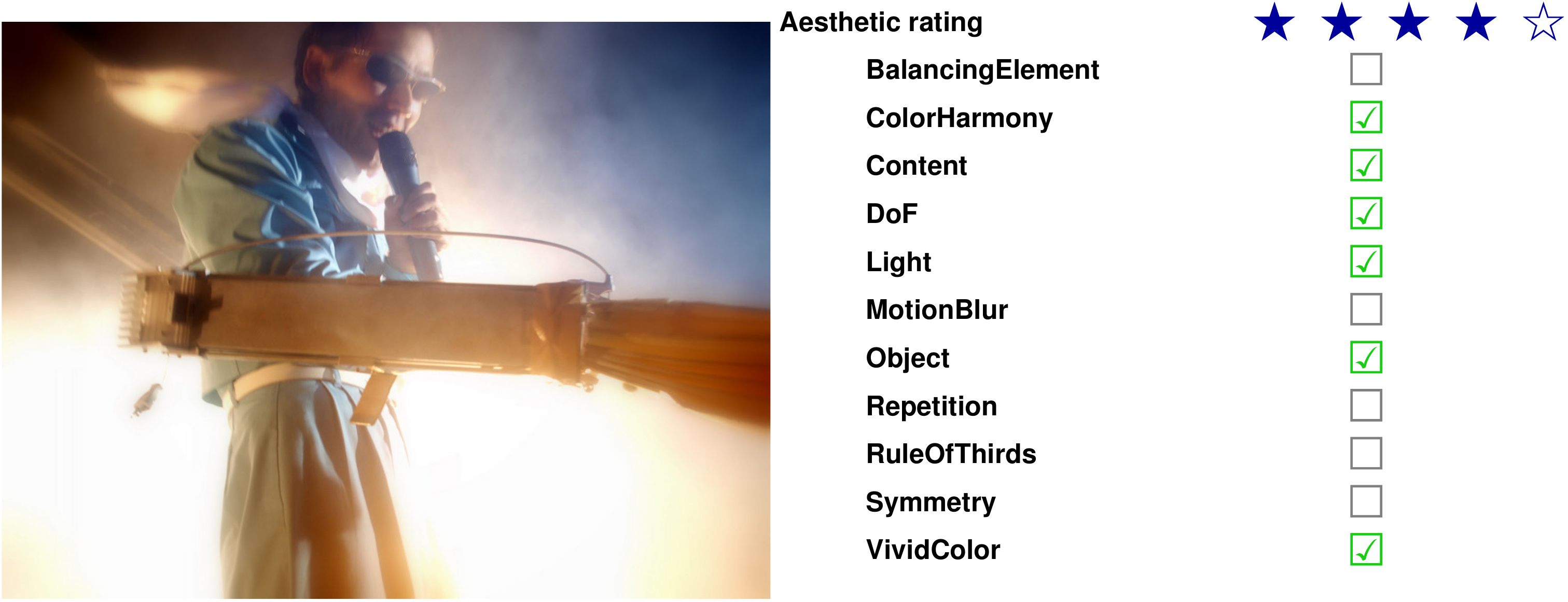}
    \end{minipage}

    \begin{minipage}{0.5\textwidth}
        \centering
        \includegraphics[width=0.99\linewidth]{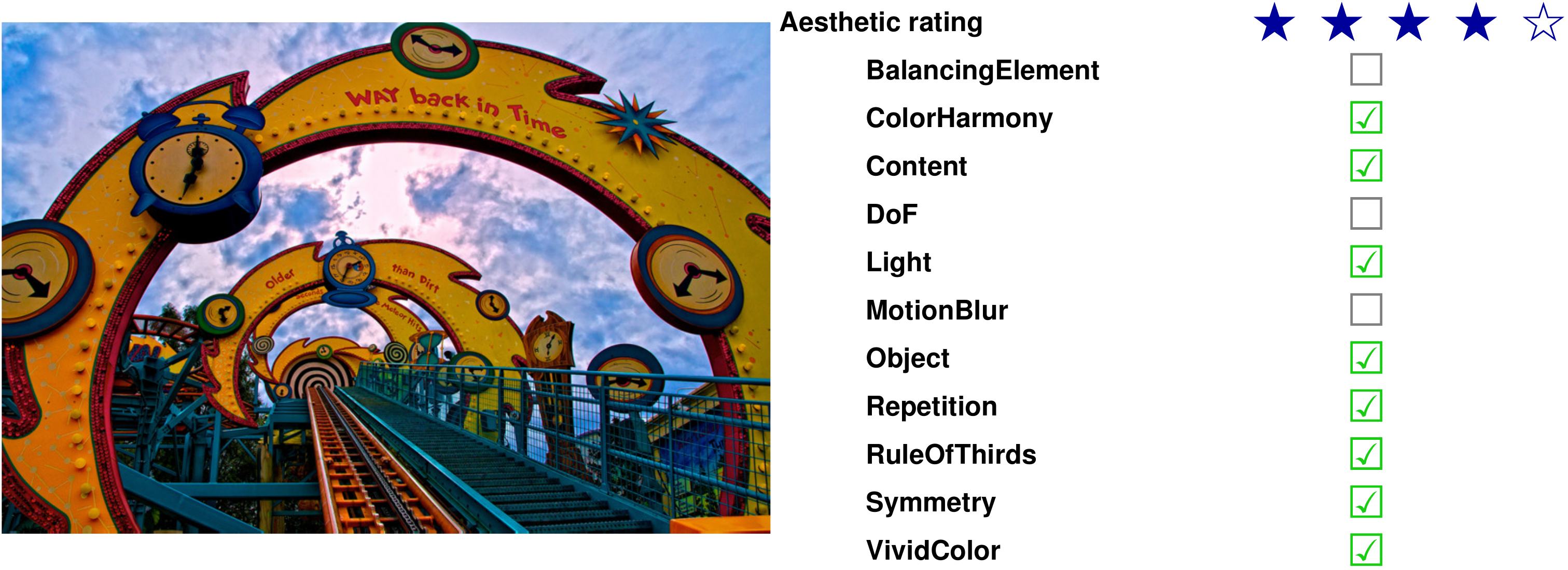}
    \end{minipage}%
    \begin{minipage}{0.5\textwidth}
        \centering
        \includegraphics[width=0.99\linewidth]{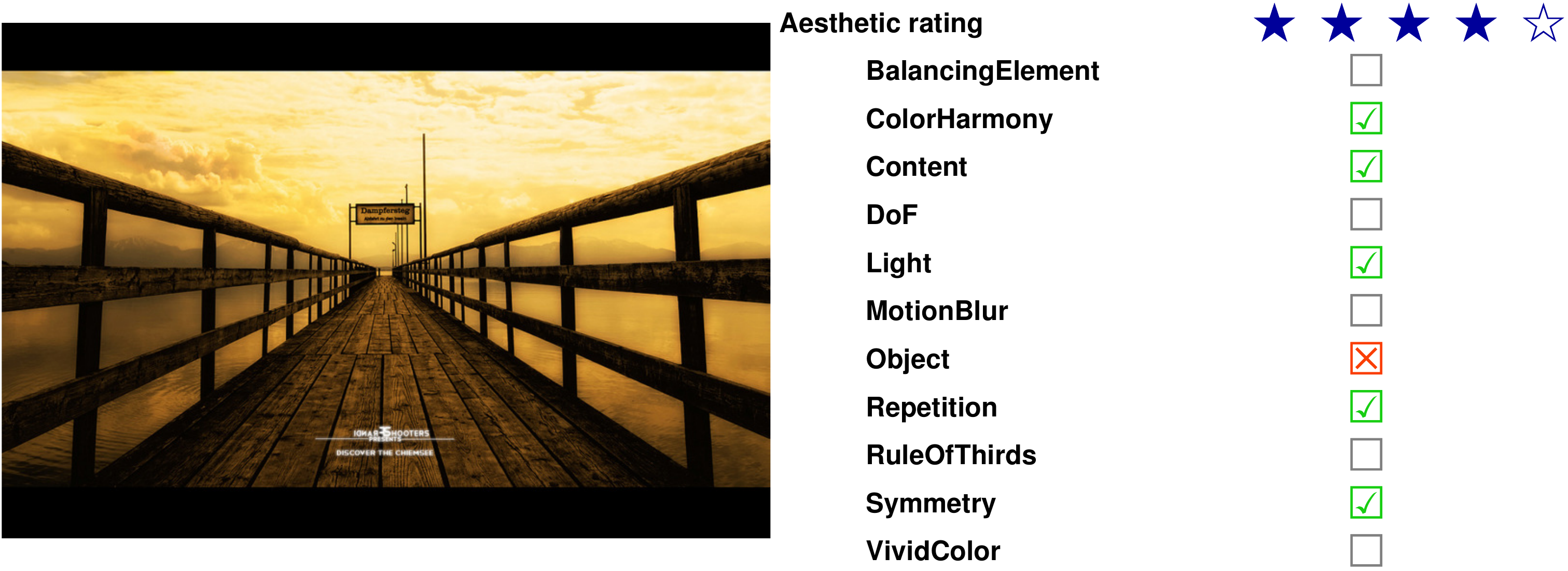}
    \end{minipage}

    \begin{minipage}{0.5\textwidth}
        \centering
        \includegraphics[width=0.99\linewidth]{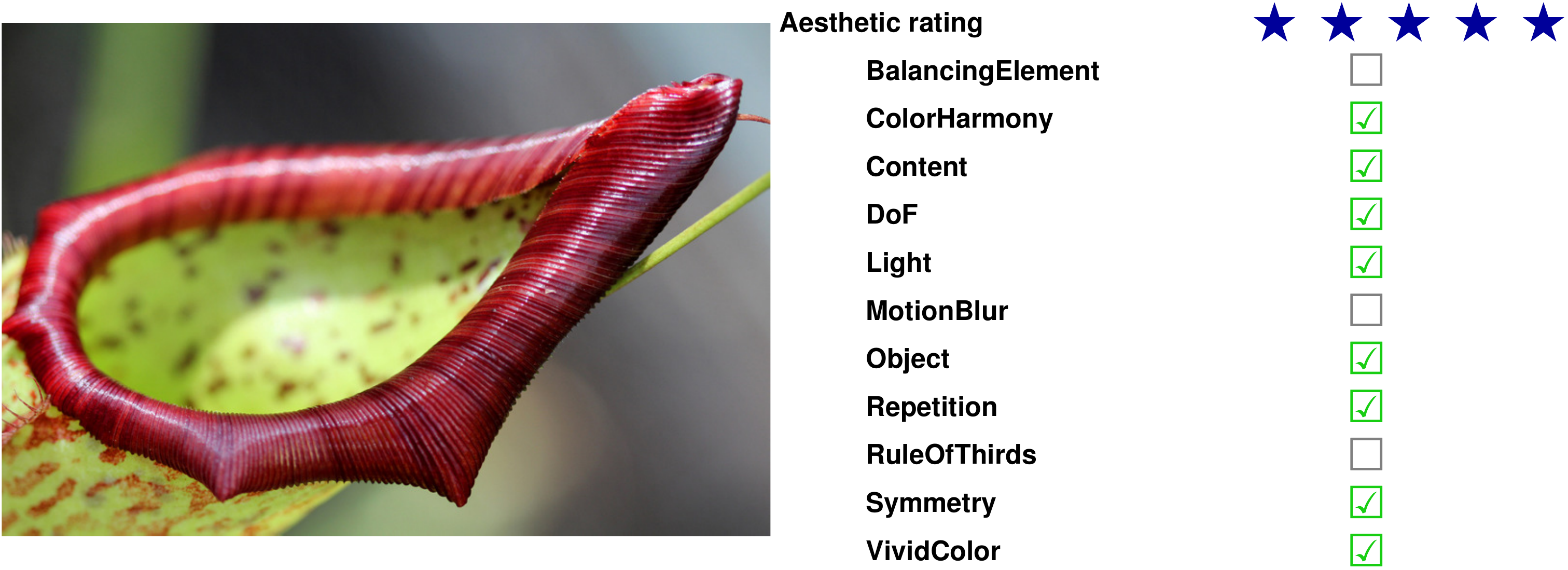}
    \end{minipage}%
    \begin{minipage}{0.5\textwidth}
        \centering
        \includegraphics[width=0.99\linewidth]{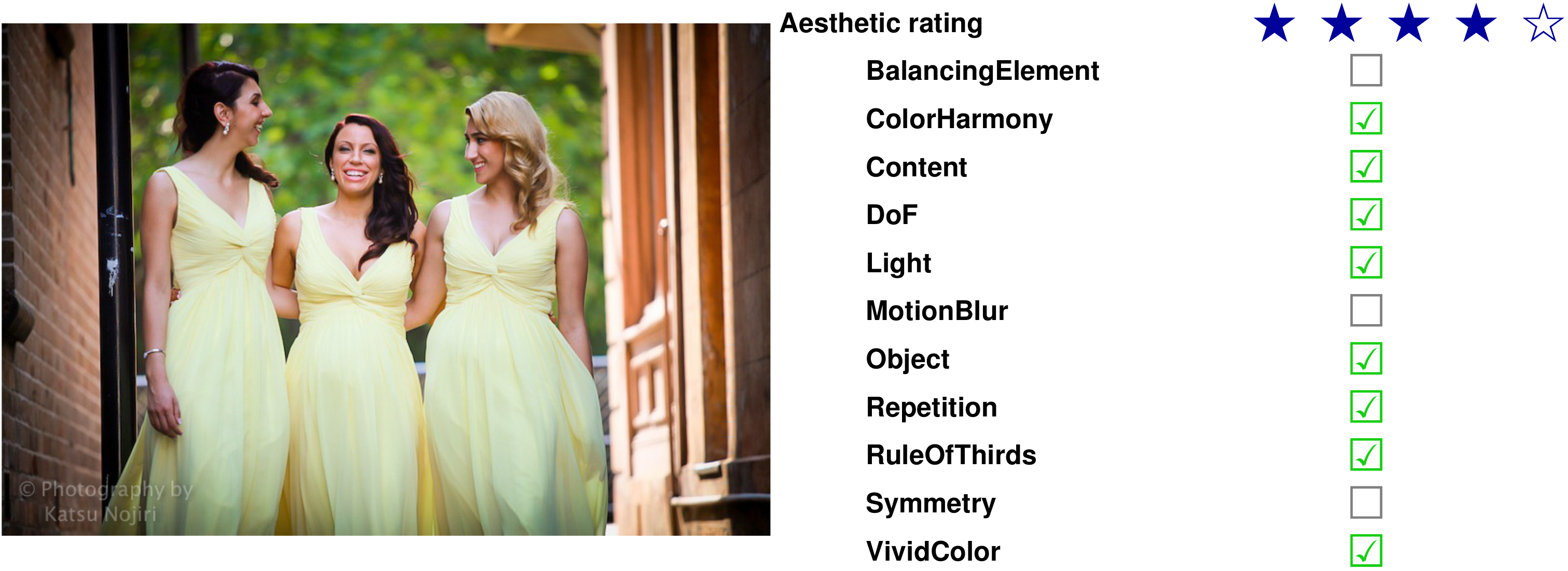}
    \end{minipage}
    \vspace{10mm}
    \caption{Some images outside our database with high estimated scores.}
    \label{fig:highImages}
\end{figure*}

\begin{figure*}[!htb]
    \centering
    \begin{minipage}{.5\textwidth}
        \centering
        \includegraphics[width=0.99\linewidth]{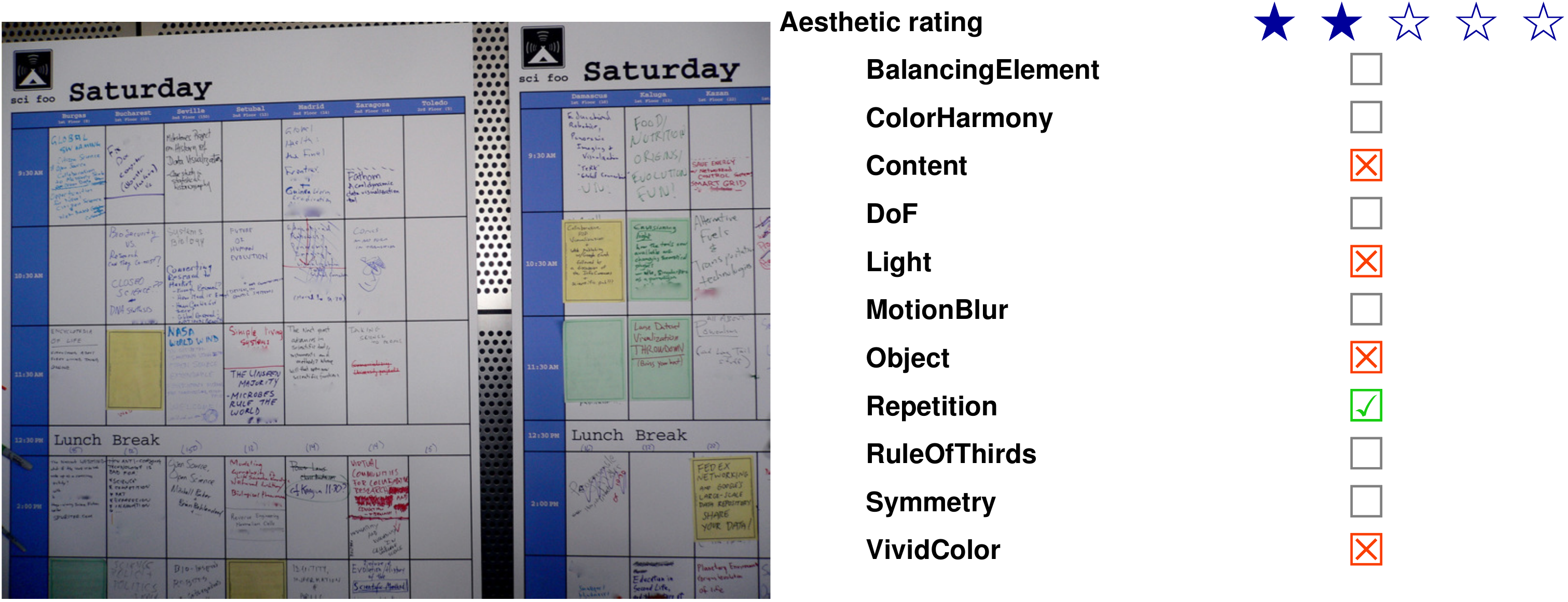}
    \end{minipage}%
    \begin{minipage}{0.5\textwidth}
        \centering
        \includegraphics[width=0.99\linewidth]{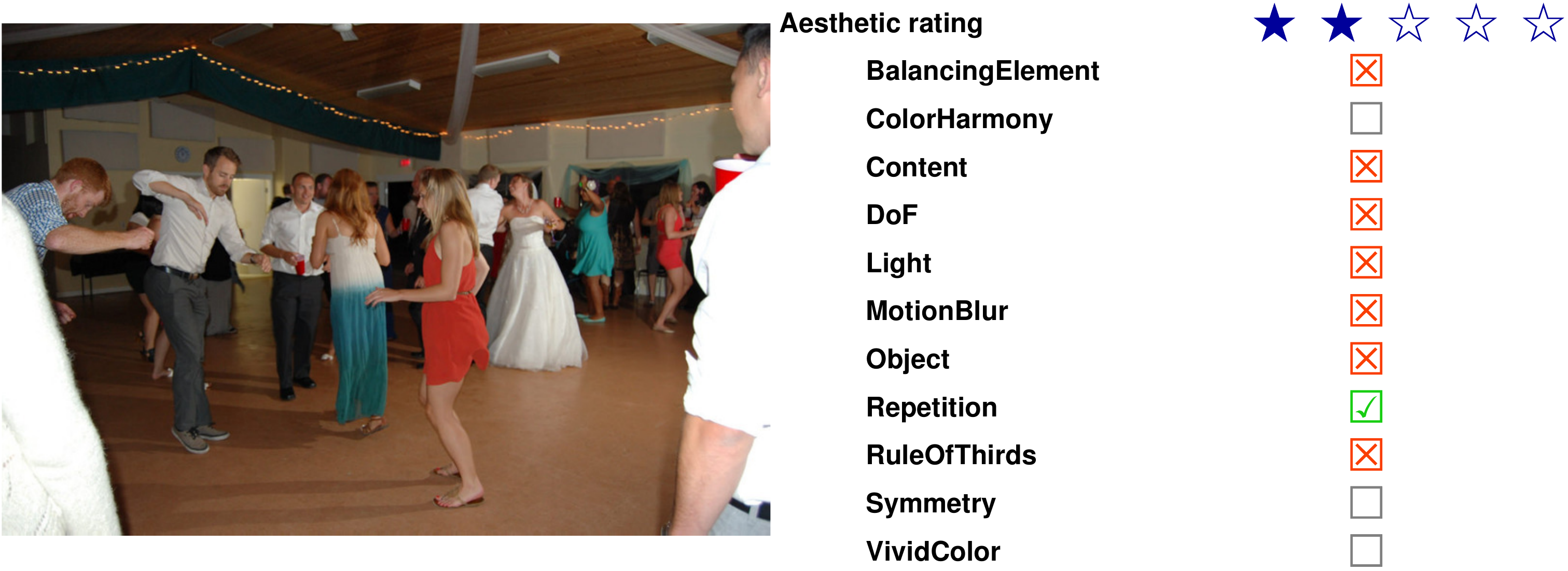}
    \end{minipage}

    \begin{minipage}{0.5\textwidth}
        \centering
        \includegraphics[width=0.99\linewidth]{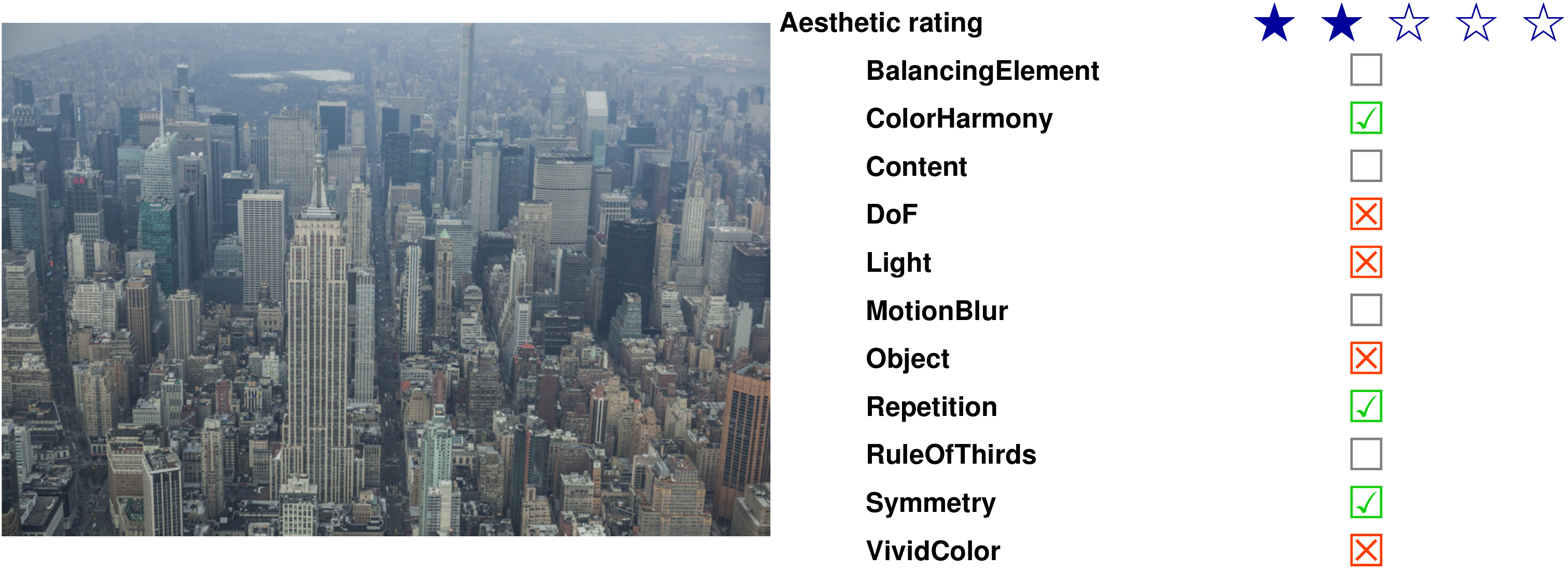}
    \end{minipage}%
    \begin{minipage}{0.5\textwidth}
        \centering
        \includegraphics[width=0.99\linewidth]{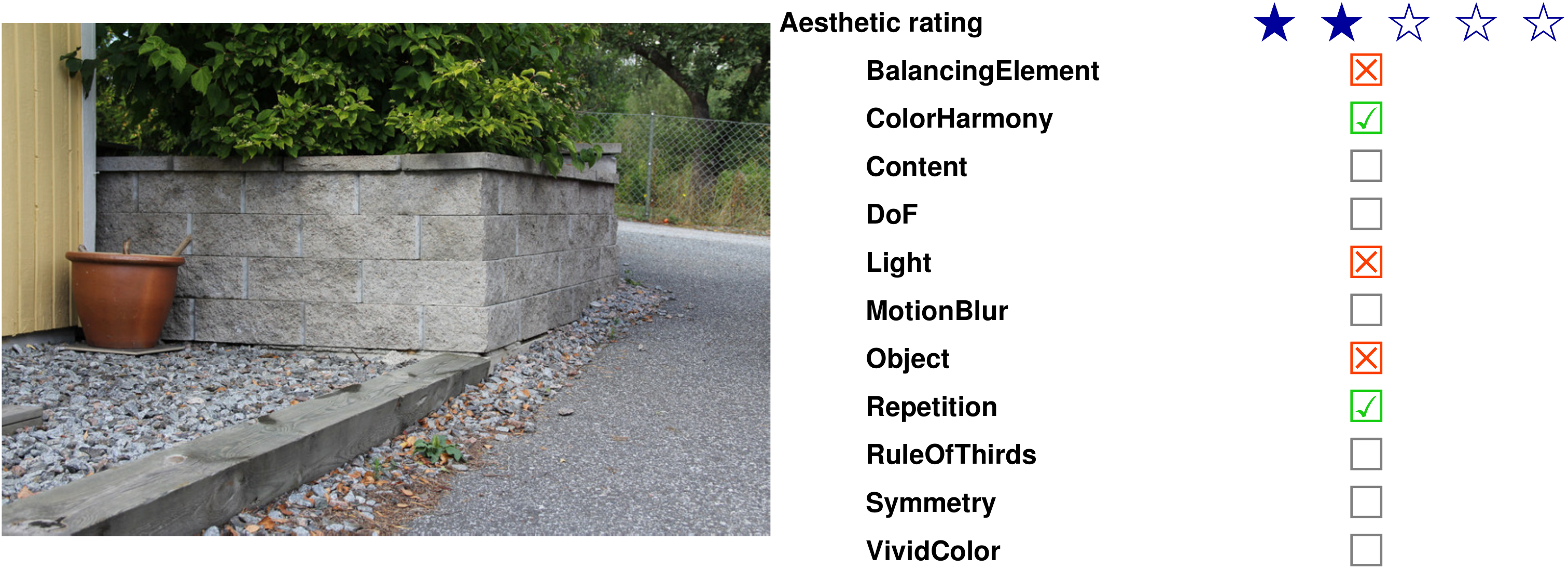}
    \end{minipage}

    \begin{minipage}{0.5\textwidth}
        \centering
        \includegraphics[width=0.99\linewidth]{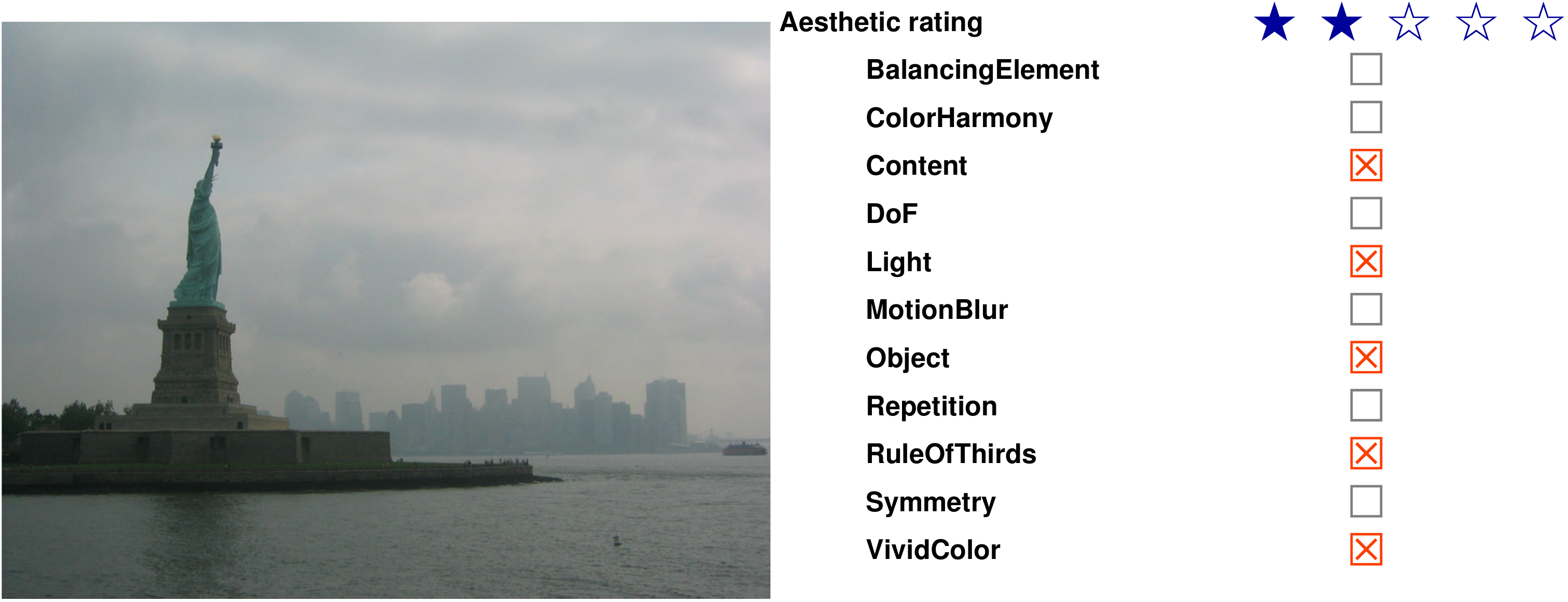}
    \end{minipage}%
    \begin{minipage}{0.5\textwidth}
        \centering
        \includegraphics[width=0.99\linewidth]{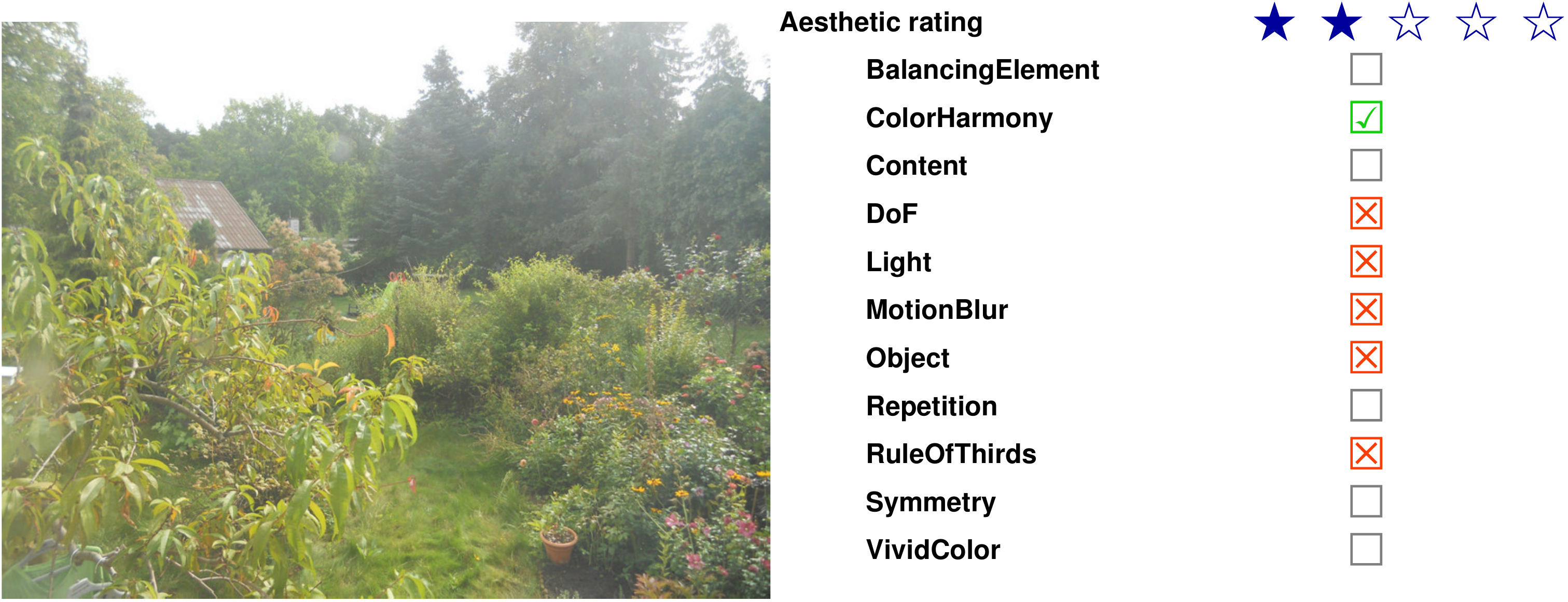}
    \end{minipage}

    \begin{minipage}{0.5\textwidth}
        \centering
        \includegraphics[width=0.99\linewidth]{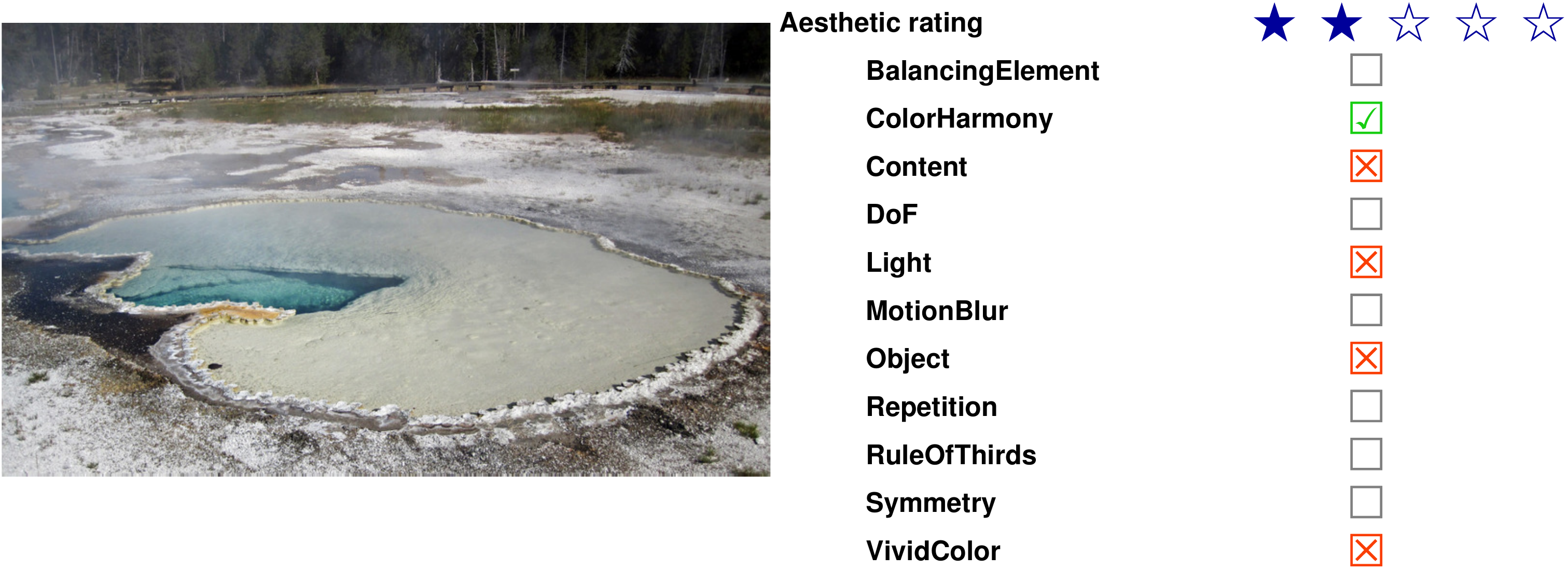}
    \end{minipage}%
    \begin{minipage}{0.5\textwidth}
        \centering
        \includegraphics[width=0.99\linewidth]{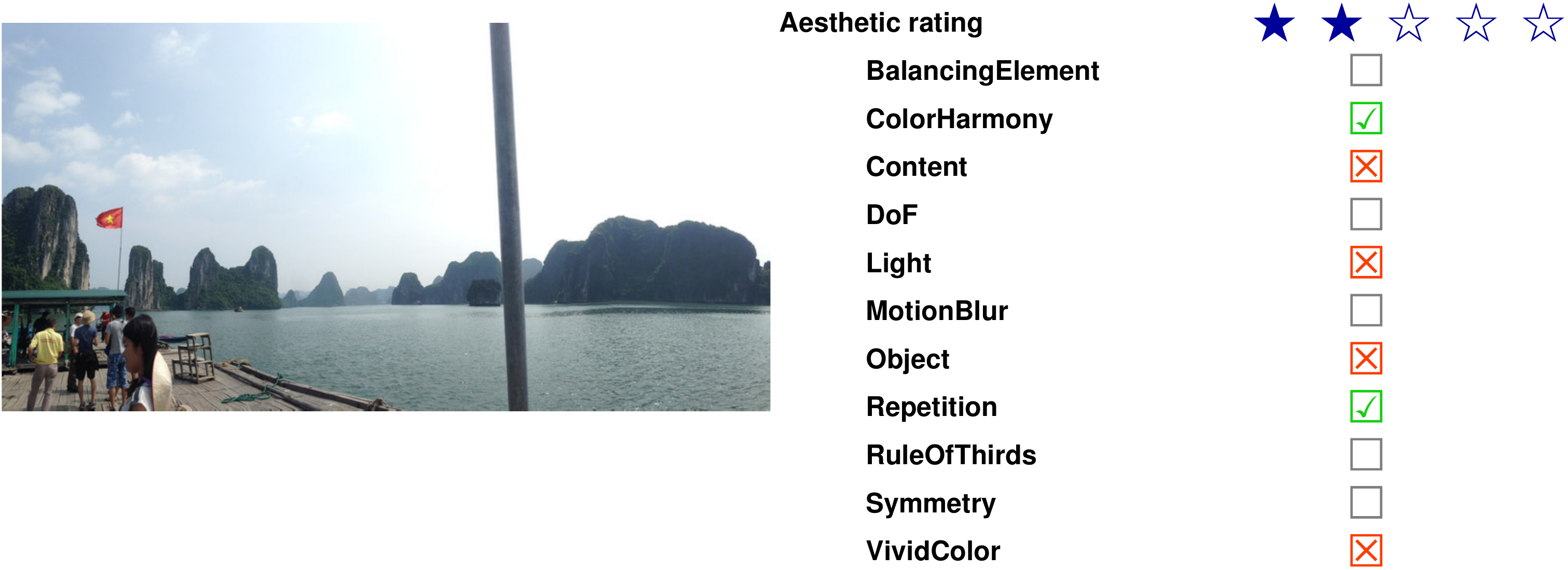}
    \end{minipage}

    \begin{minipage}{0.5\textwidth}
        \centering
        \includegraphics[width=0.99\linewidth]{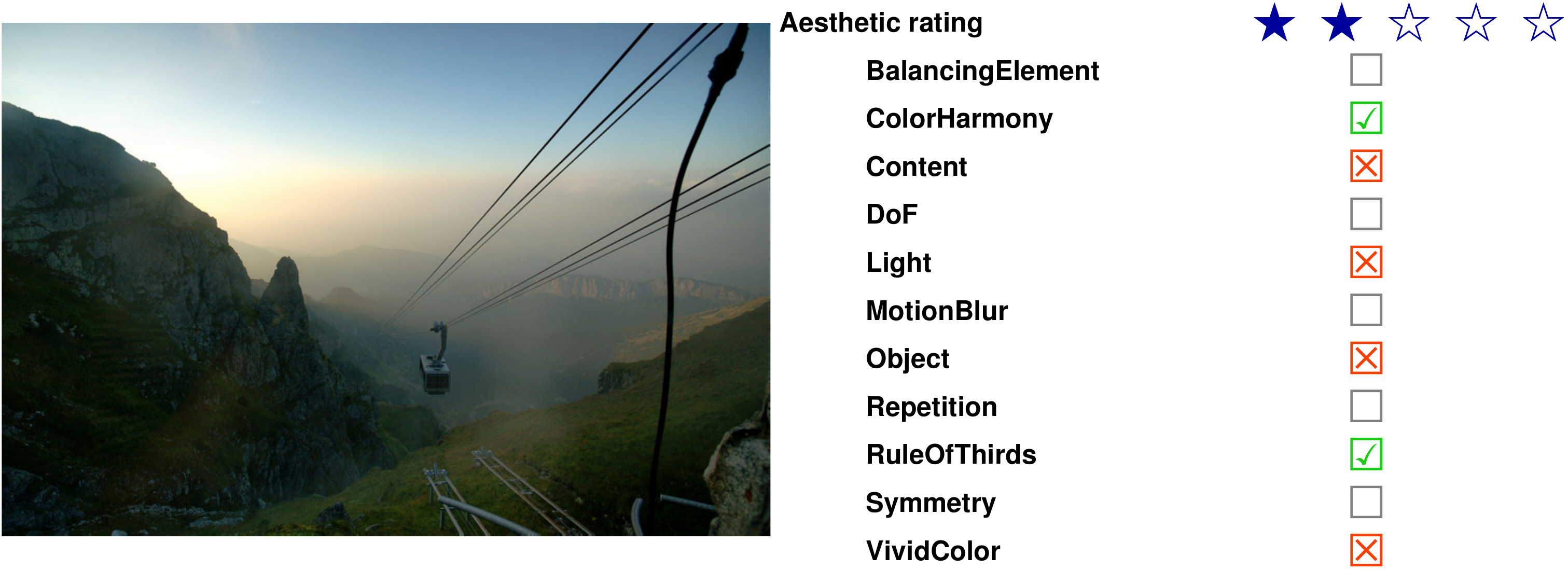}
    \end{minipage}%
    \begin{minipage}{0.5\textwidth}
        \centering
        \includegraphics[width=0.99\linewidth]{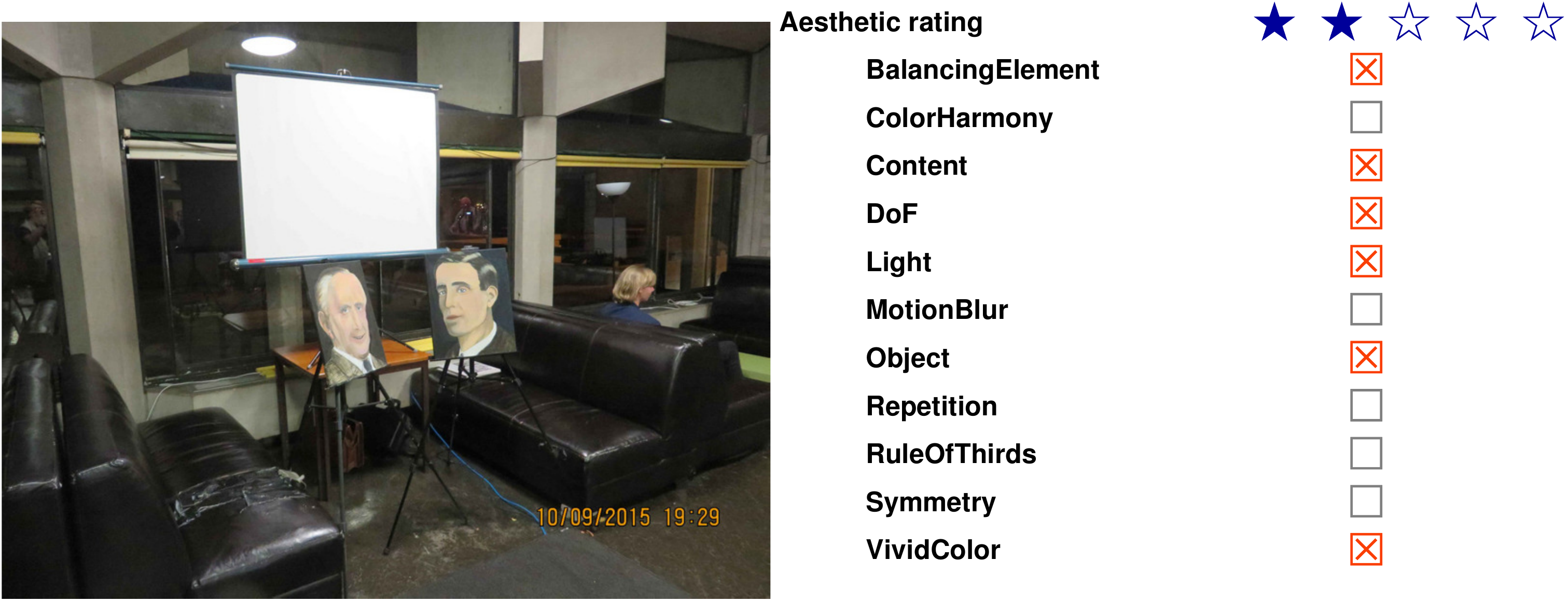}
    \end{minipage}

    \begin{minipage}{0.5\textwidth}
        \centering
        \includegraphics[width=0.99\linewidth]{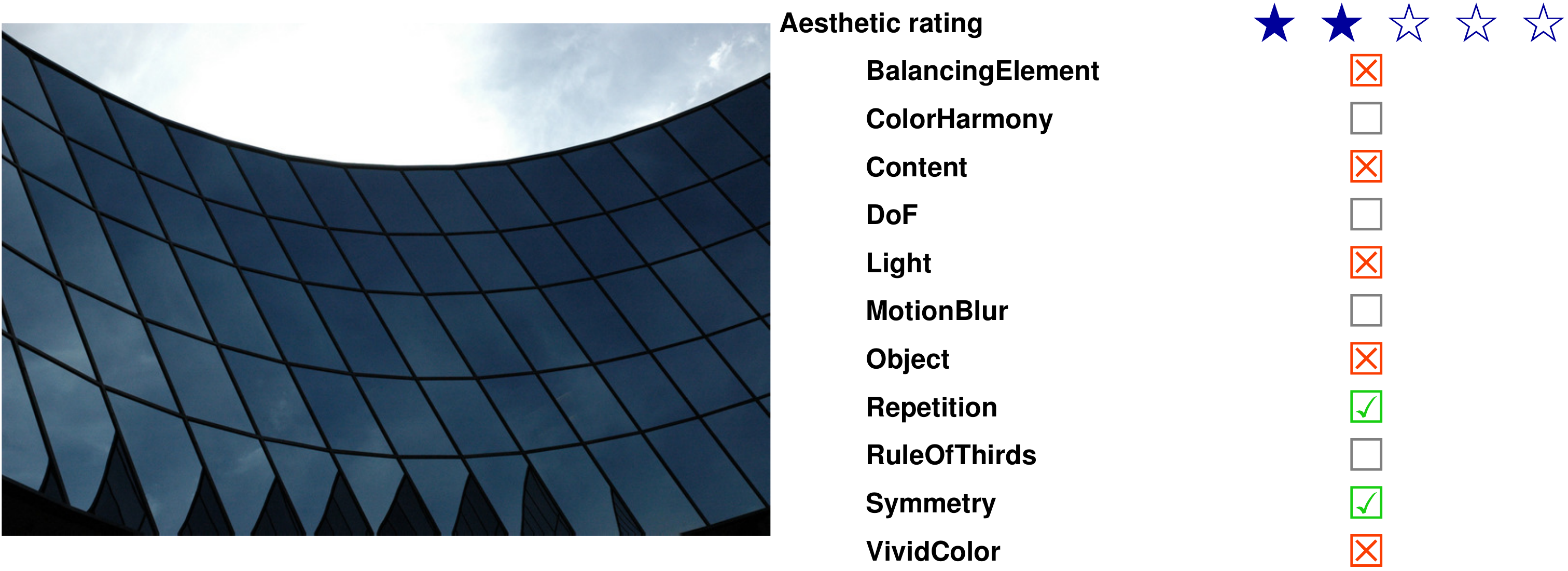}
    \end{minipage}%
    \begin{minipage}{0.5\textwidth}
        \centering
        \includegraphics[width=0.99\linewidth]{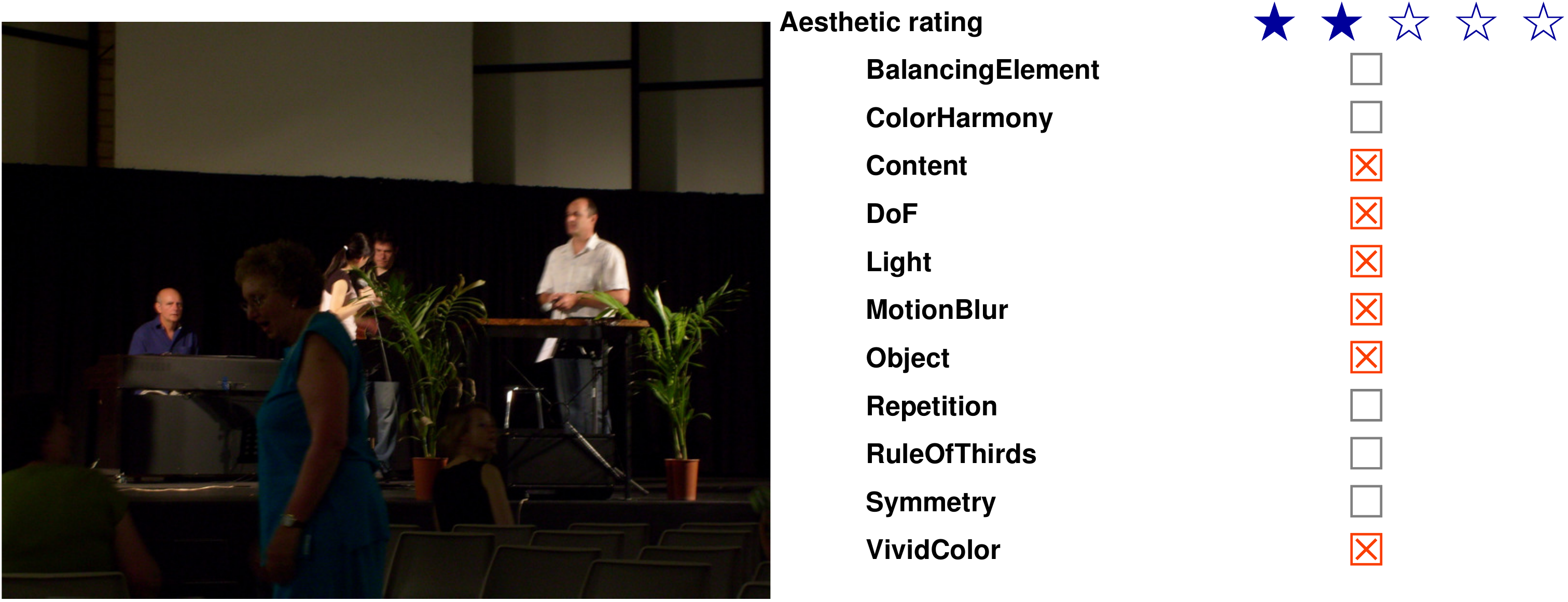}
    \end{minipage}

    \begin{minipage}{0.5\textwidth}
        \centering
        \includegraphics[width=0.99\linewidth]{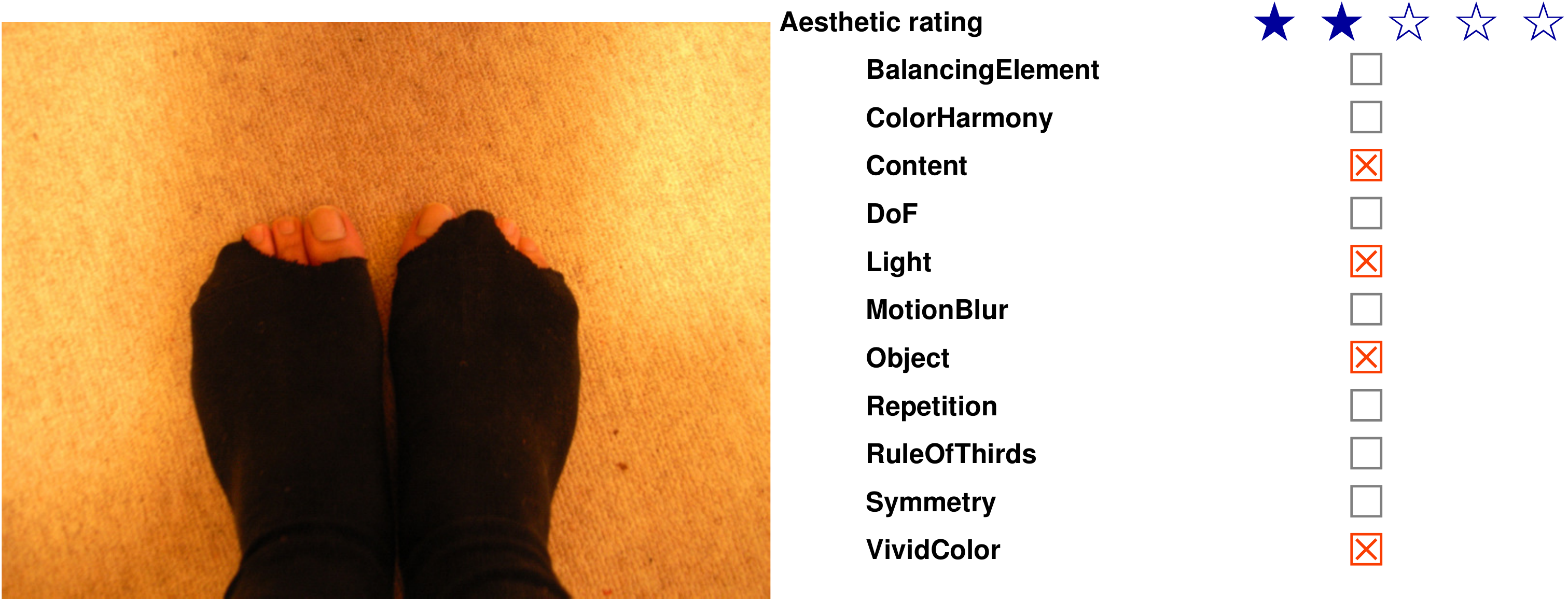}
    \end{minipage}%
    \begin{minipage}{0.5\textwidth}
        \centering
        \includegraphics[width=0.99\linewidth]{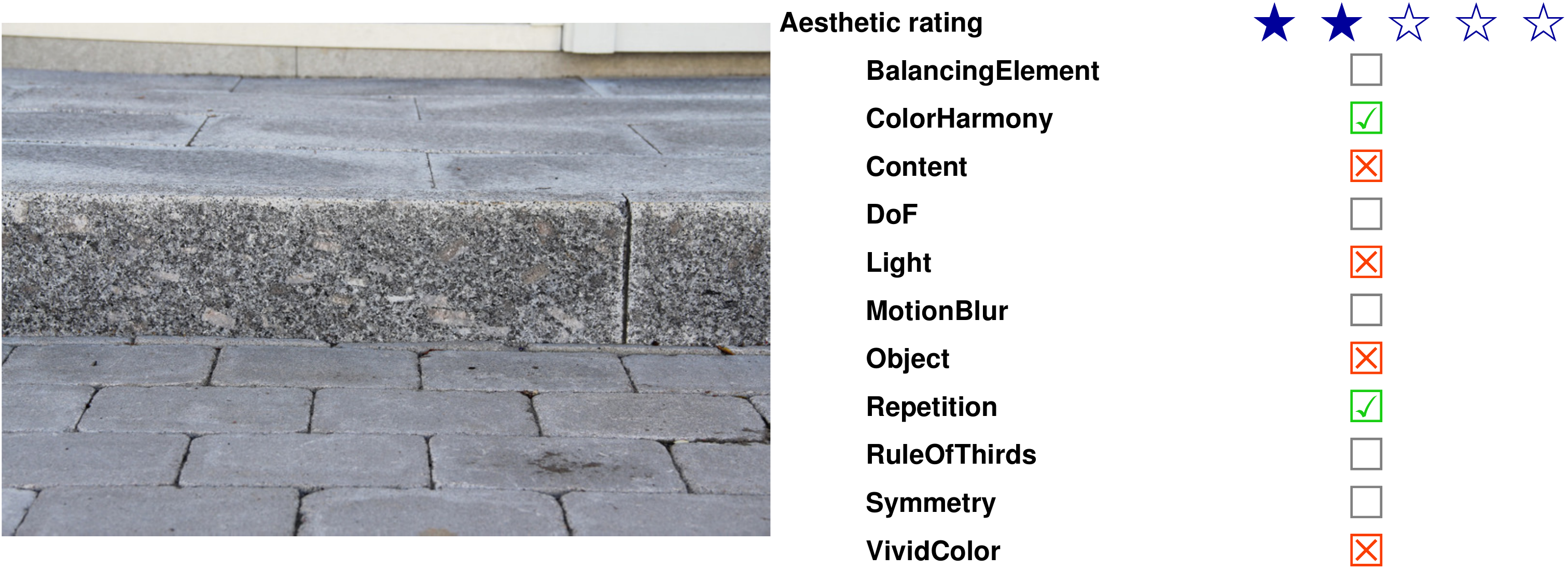}
    \end{minipage}
    \vspace{10mm}

    \caption{Some images outside our database with low estimated scores.}
    \label{fig:lowImages}
\end{figure*}

\begin{figure*}[!htb]
    \centering
    \begin{minipage}{.5\textwidth}
        \centering
        \includegraphics[width=0.99\linewidth]{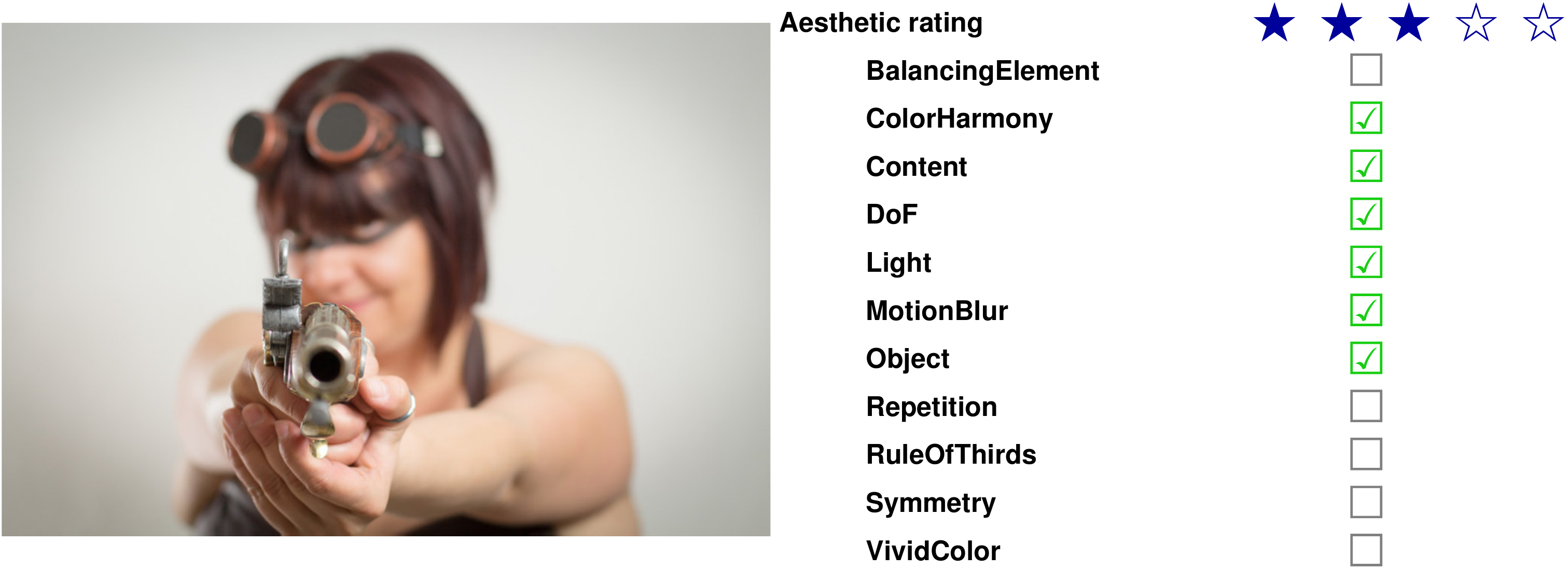}
    \end{minipage}%
    \begin{minipage}{0.5\textwidth}
        \centering
        \includegraphics[width=0.99\linewidth]{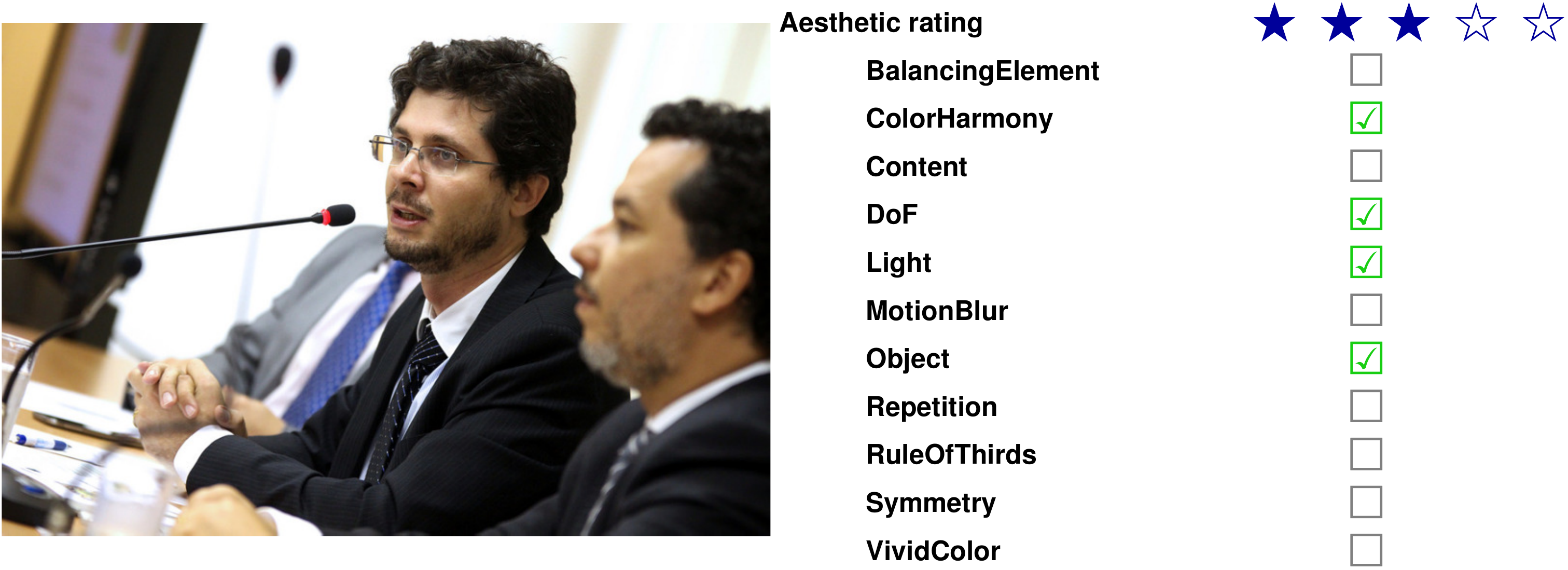}
    \end{minipage}

    \begin{minipage}{0.5\textwidth}
        \centering
        \includegraphics[width=0.99\linewidth]{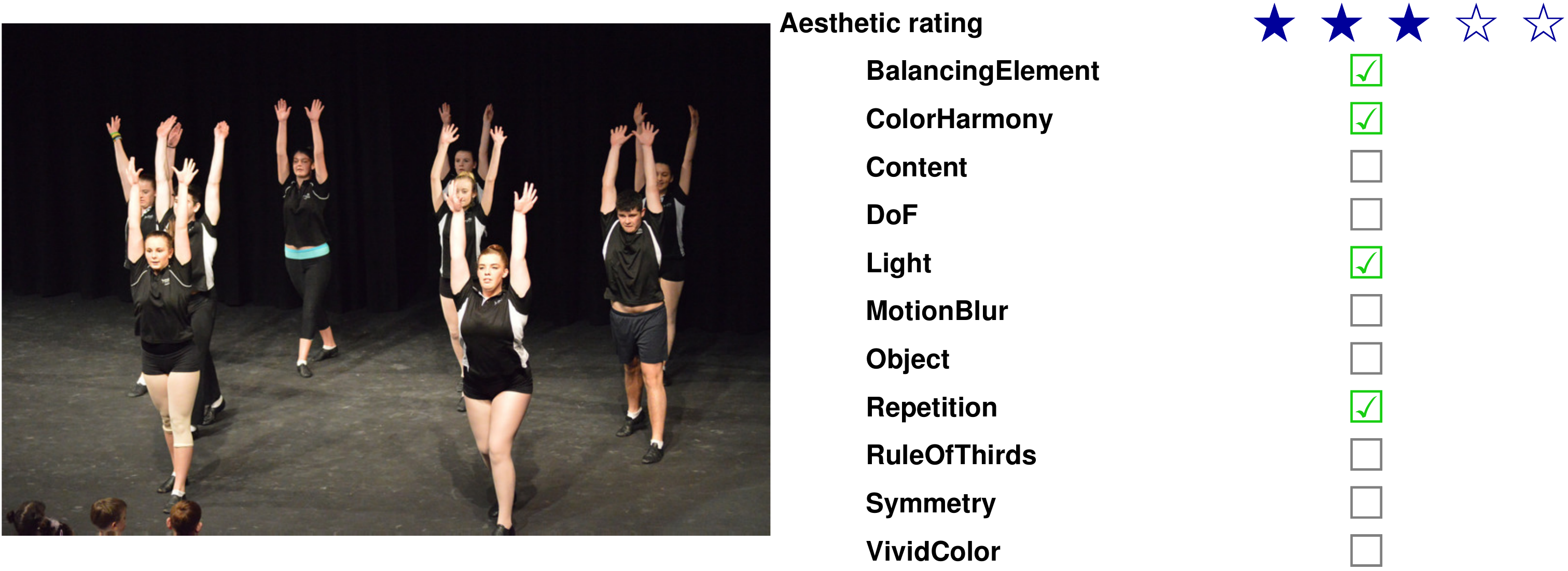}
    \end{minipage}%
    \begin{minipage}{0.5\textwidth}
        \centering
        \includegraphics[width=0.99\linewidth]{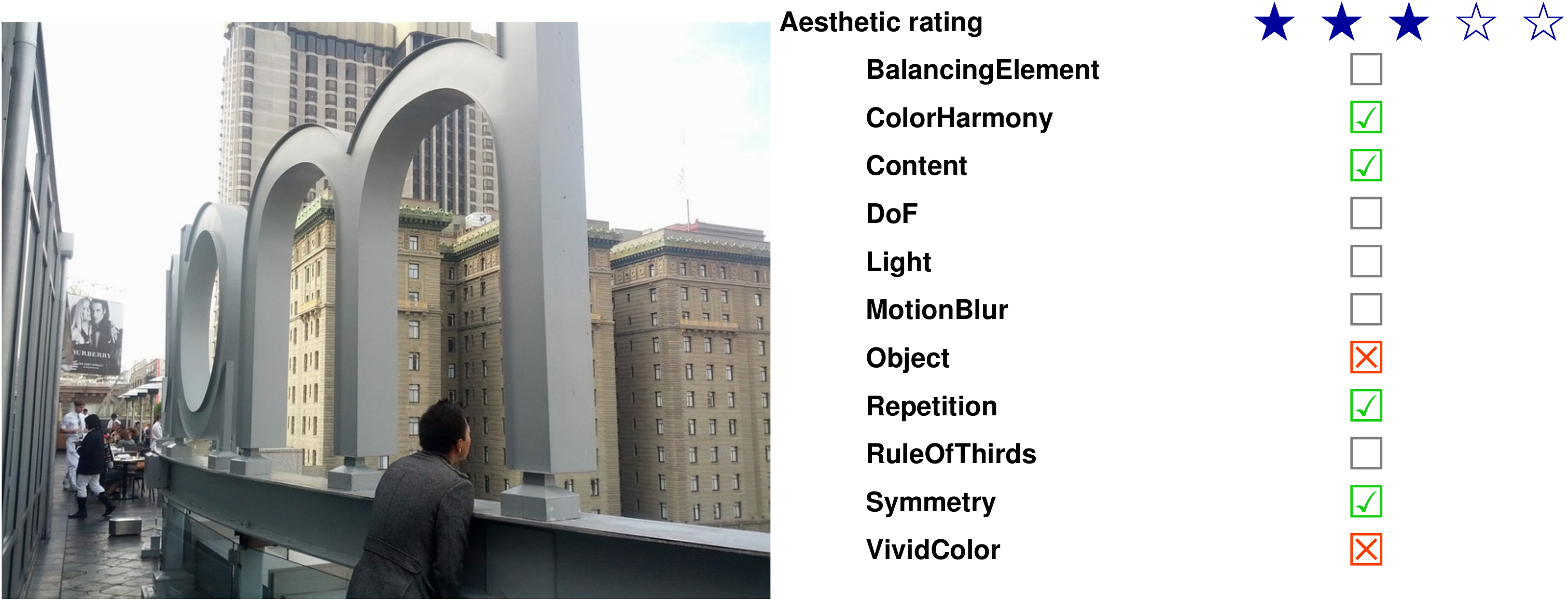}
    \end{minipage}

    \begin{minipage}{0.5\textwidth}
        \centering
        \includegraphics[width=0.99\linewidth]{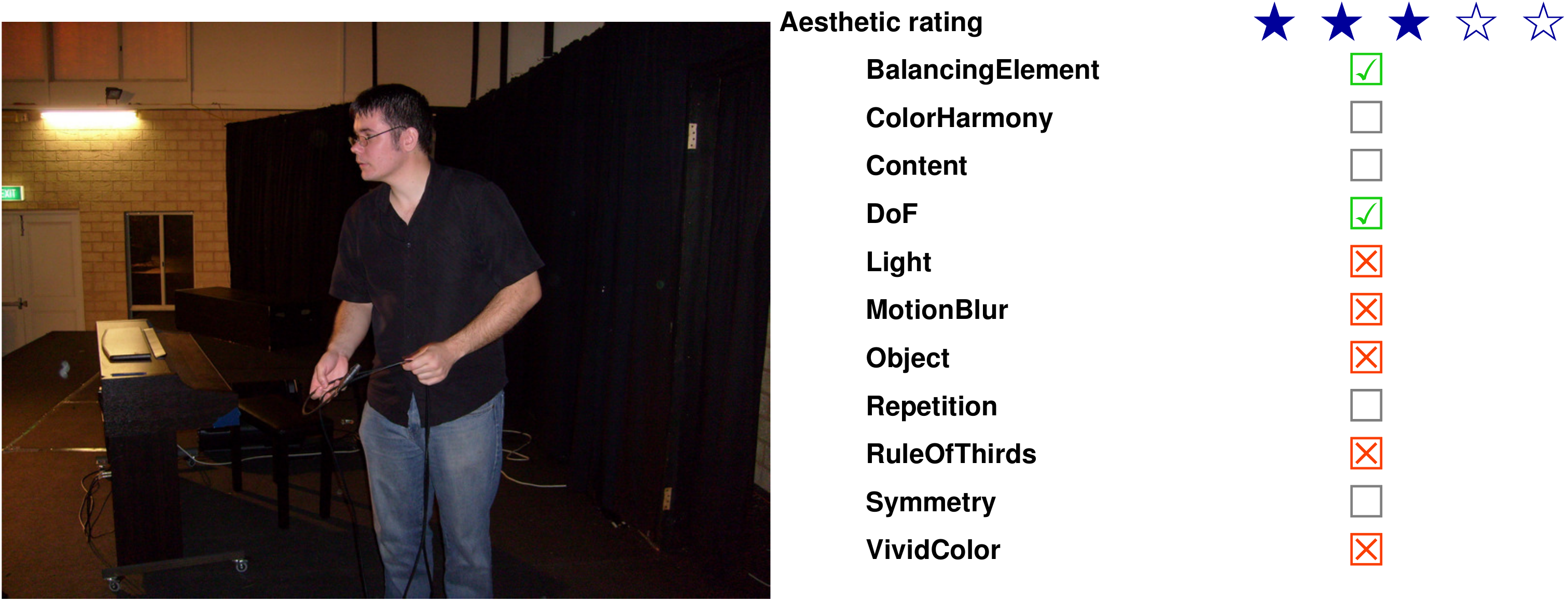}
    \end{minipage}%
    \begin{minipage}{0.5\textwidth}
        \centering
        \includegraphics[width=0.99\linewidth]{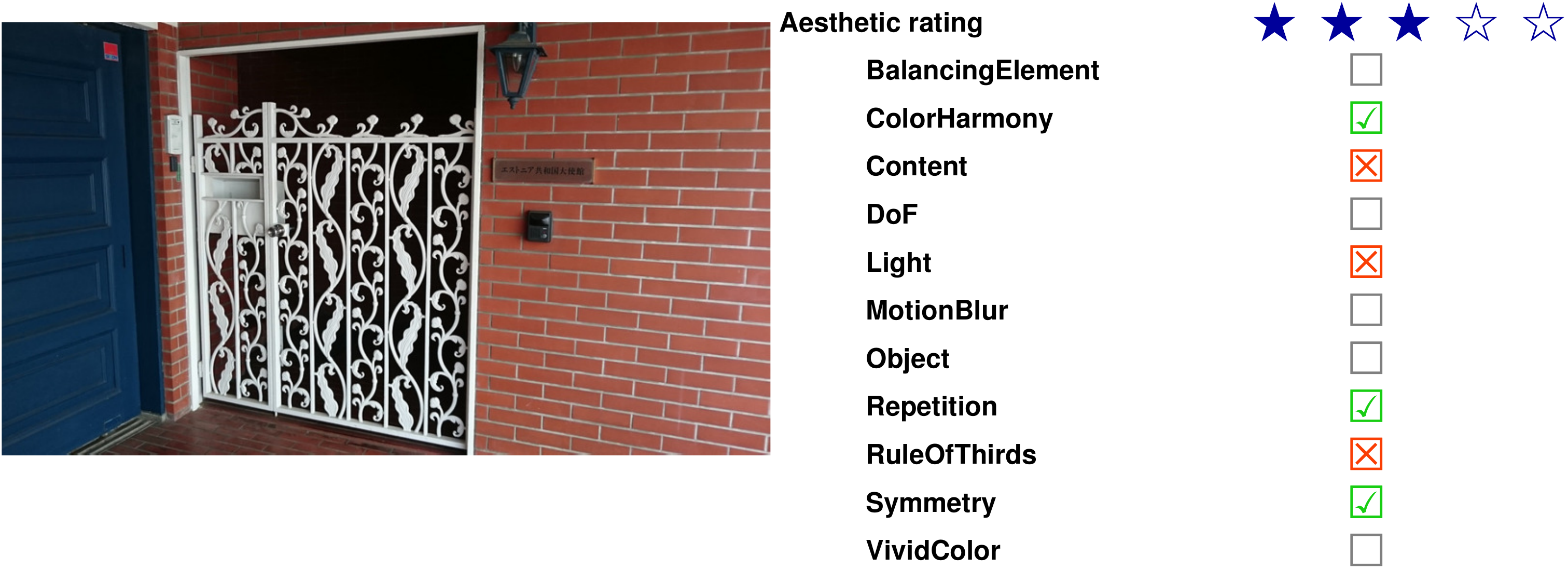}
    \end{minipage}

    \begin{minipage}{0.5\textwidth}
        \centering
        \includegraphics[width=0.99\linewidth]{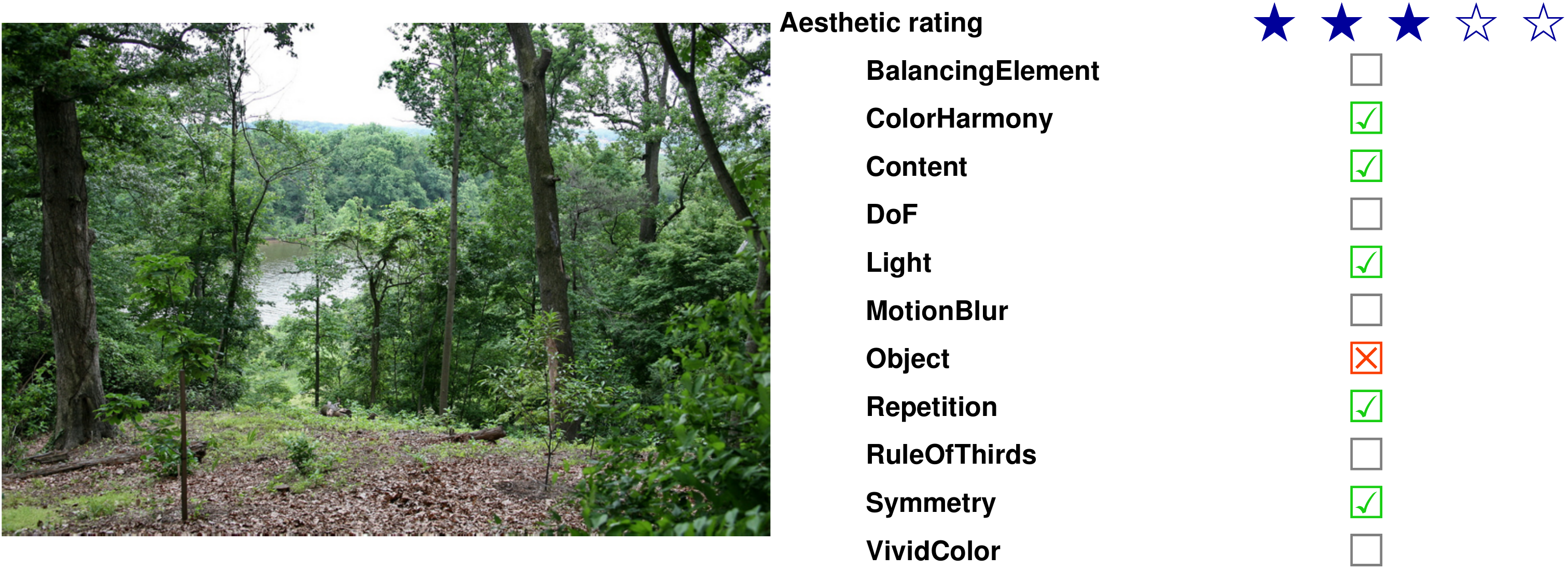}
    \end{minipage}%
    \begin{minipage}{0.5\textwidth}
        \centering
        \includegraphics[width=0.99\linewidth]{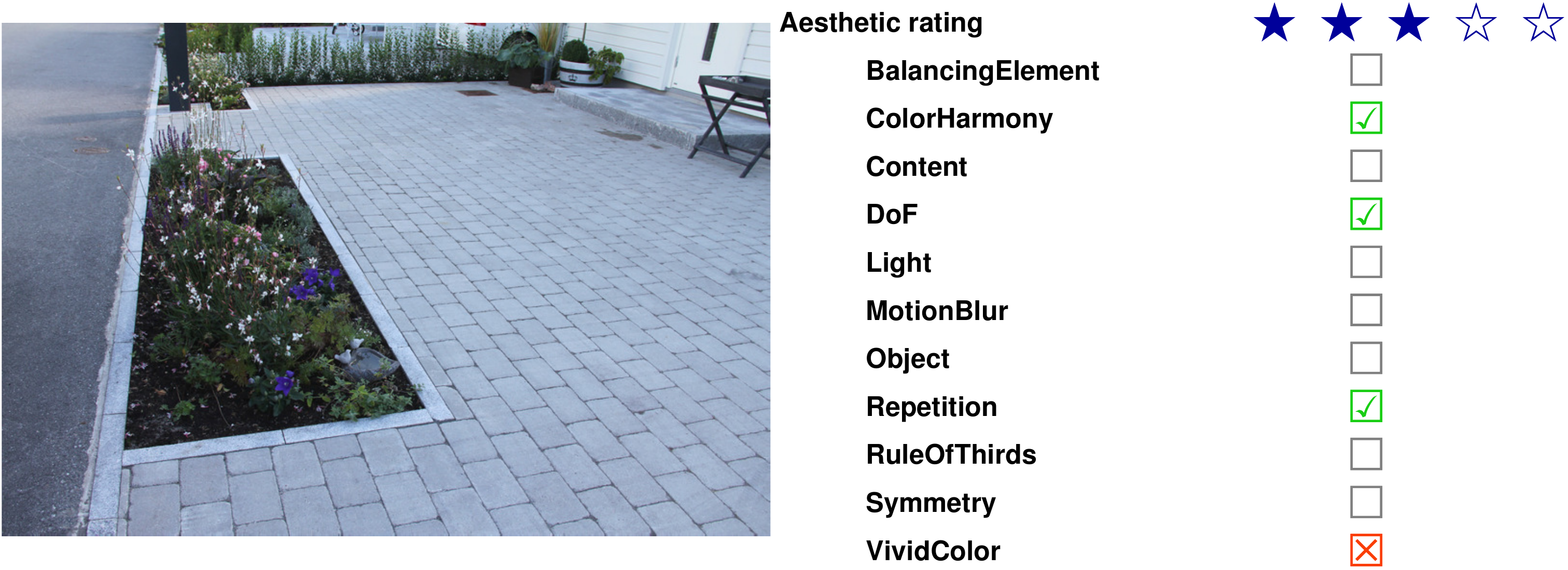}
    \end{minipage}

    \begin{minipage}{0.5\textwidth}
        \centering
        \includegraphics[width=0.99\linewidth]{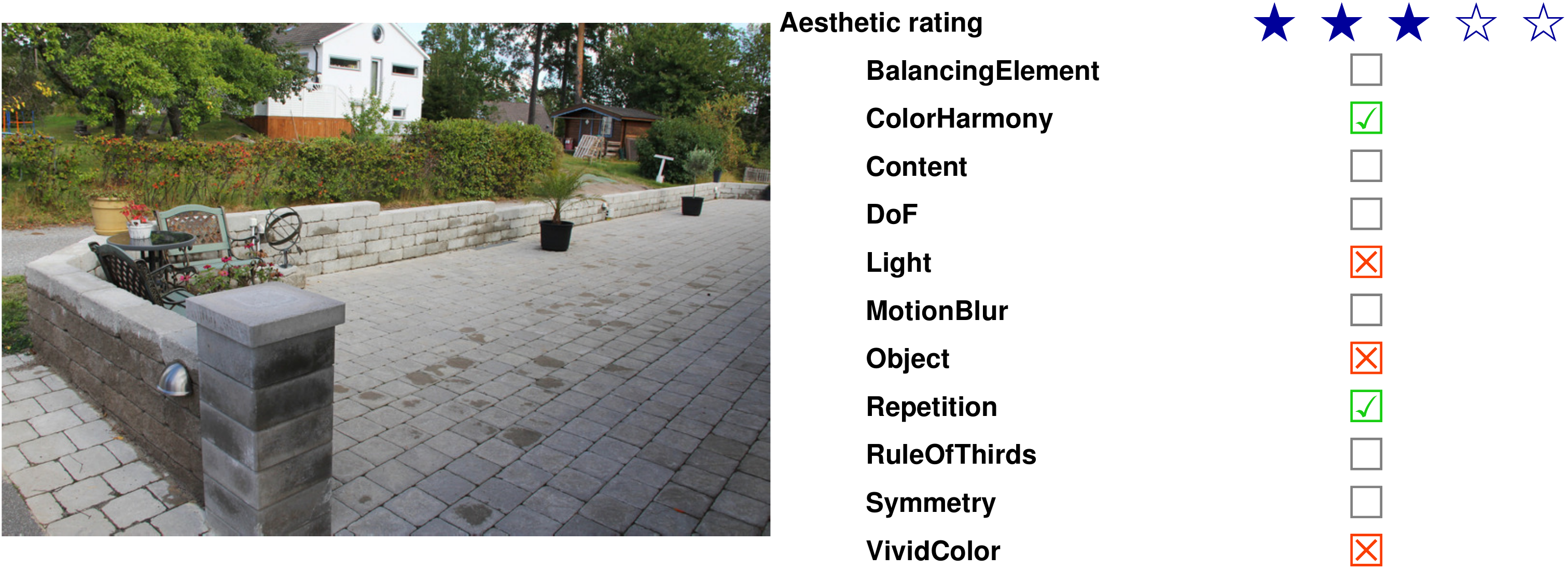}
    \end{minipage}%
    \begin{minipage}{0.5\textwidth}
        \centering
        \includegraphics[width=0.99\linewidth]{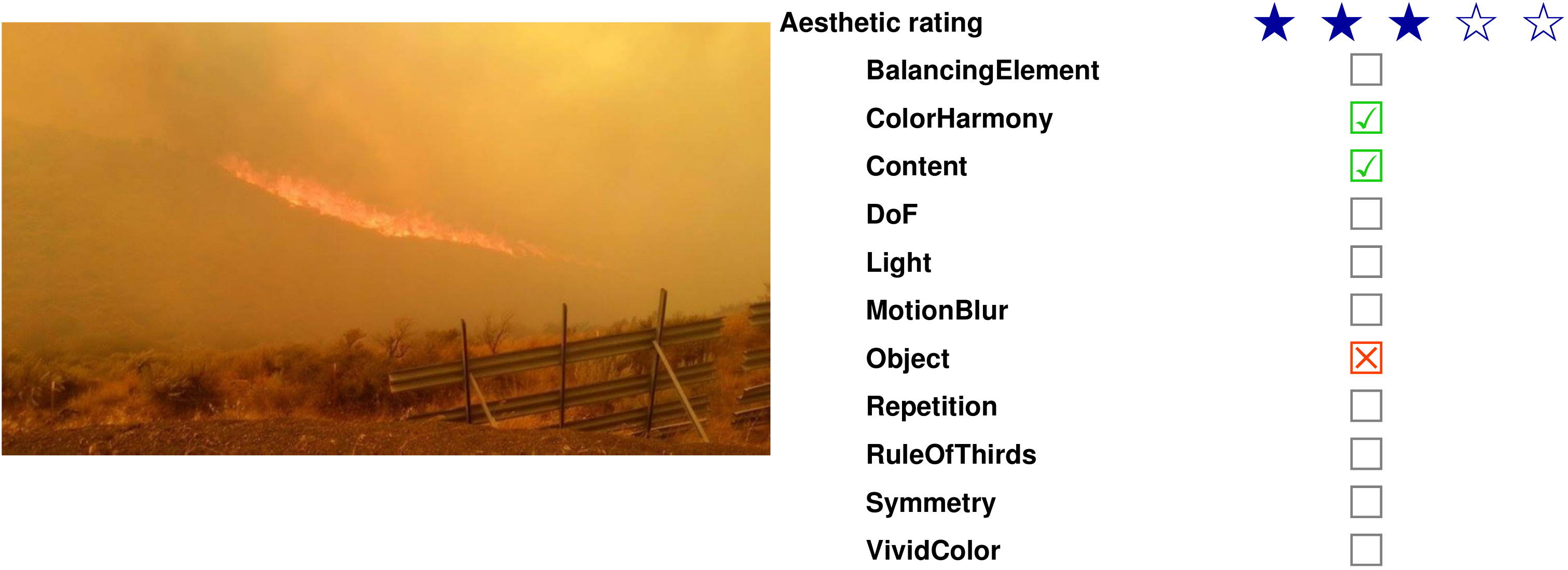}
    \end{minipage}

    \begin{minipage}{0.5\textwidth}
        \centering
        \includegraphics[width=0.99\linewidth]{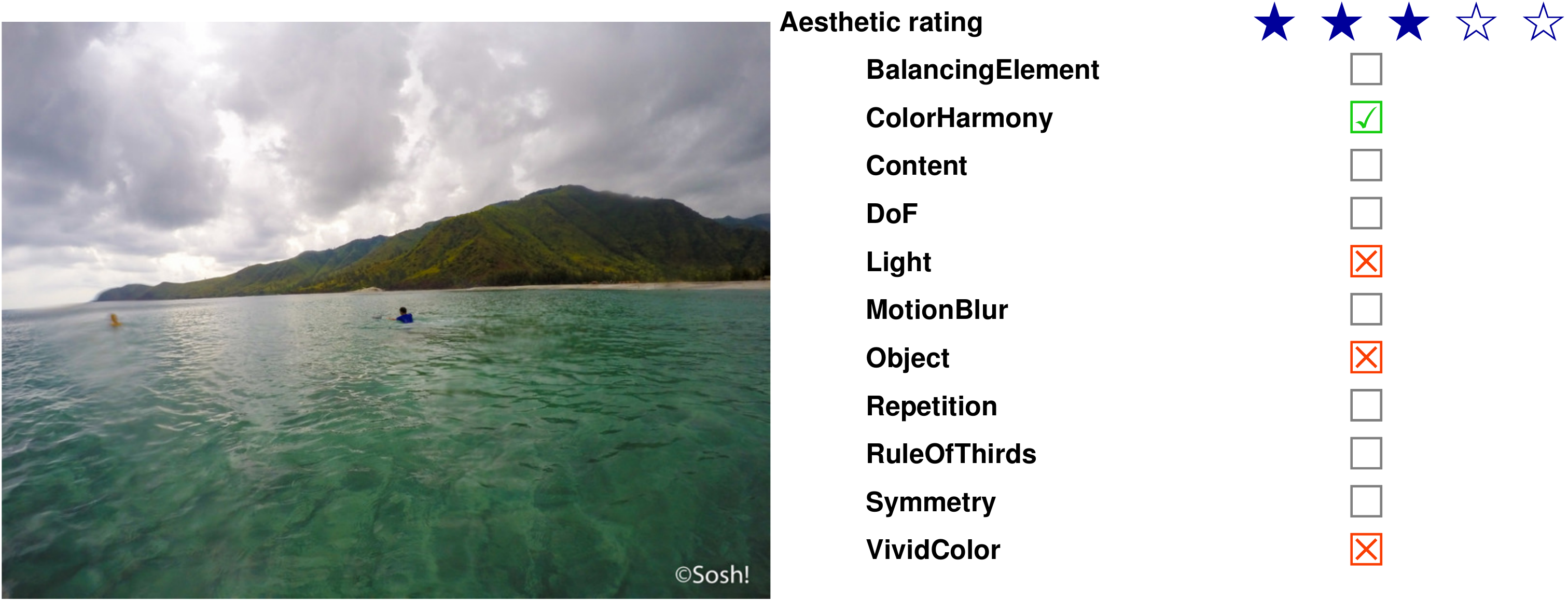}
    \end{minipage}%
    \begin{minipage}{0.5\textwidth}
        \centering
        \includegraphics[width=0.99\linewidth]{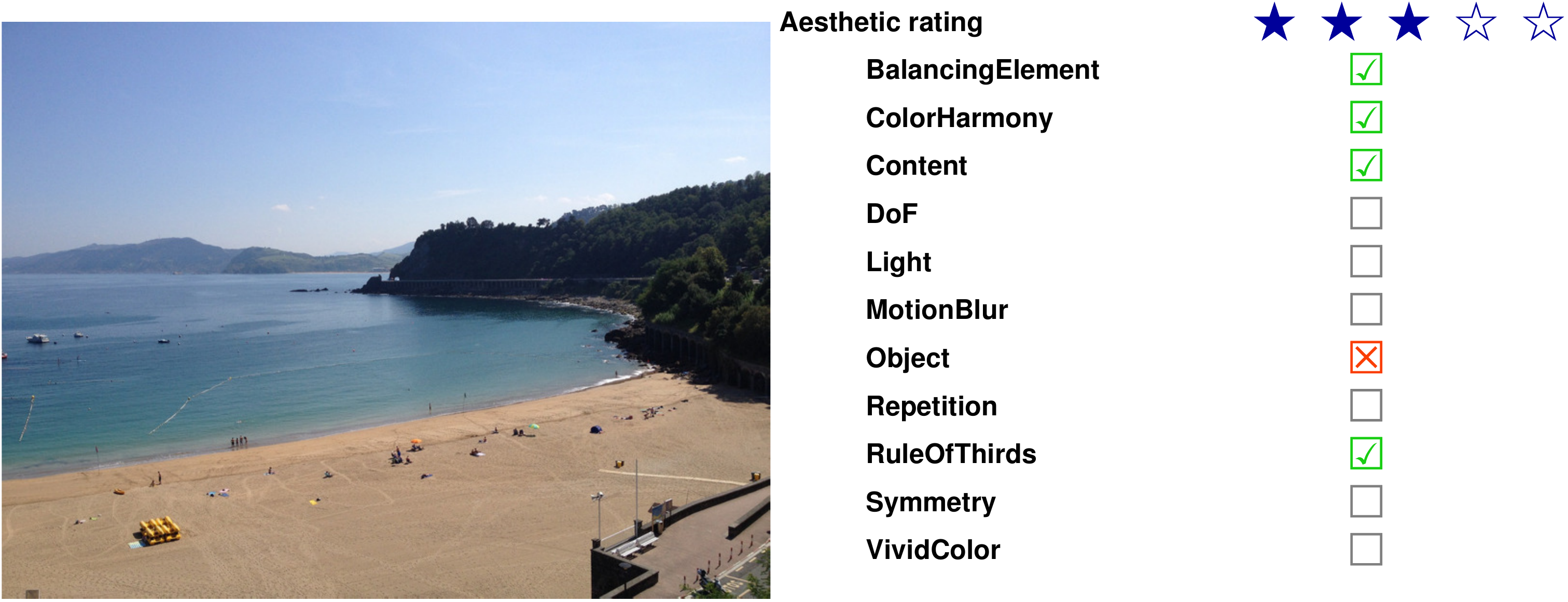}
    \end{minipage}

    \begin{minipage}{0.5\textwidth}
        \centering
        \includegraphics[width=0.99\linewidth]{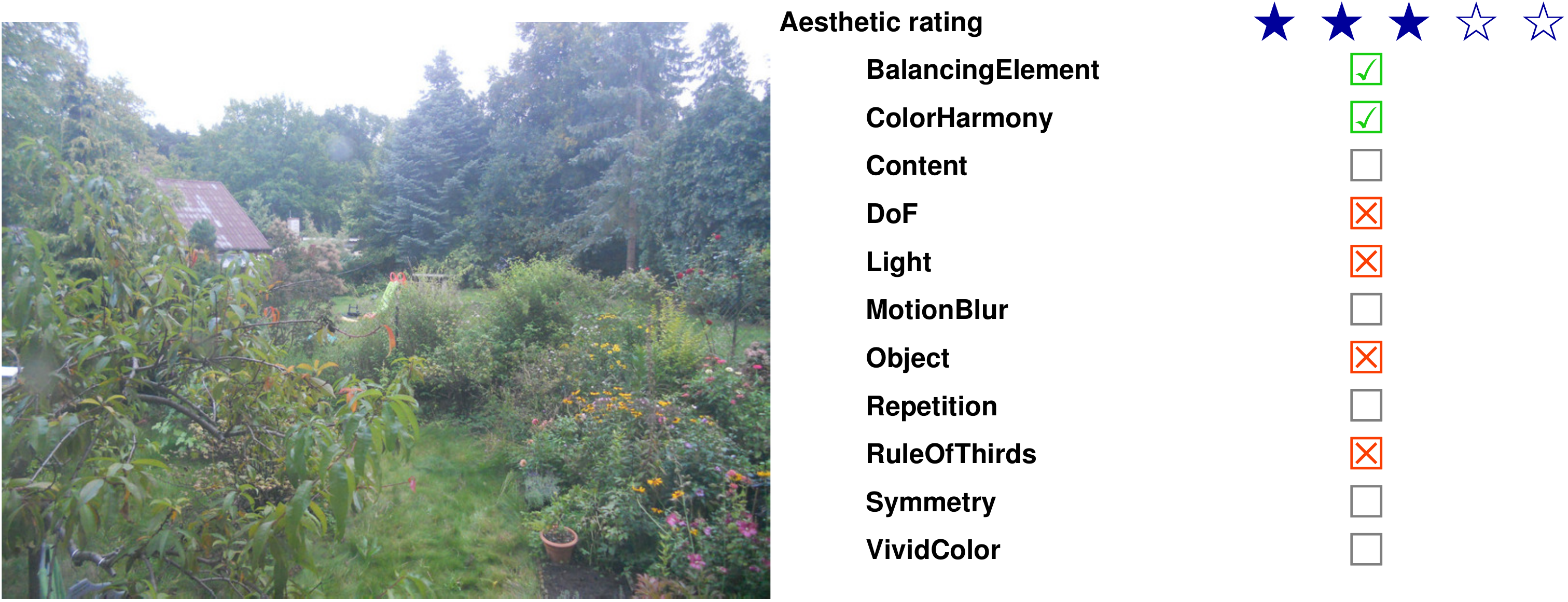}
    \end{minipage}%
    \begin{minipage}{0.5\textwidth}
        \centering
        \includegraphics[width=0.99\linewidth]{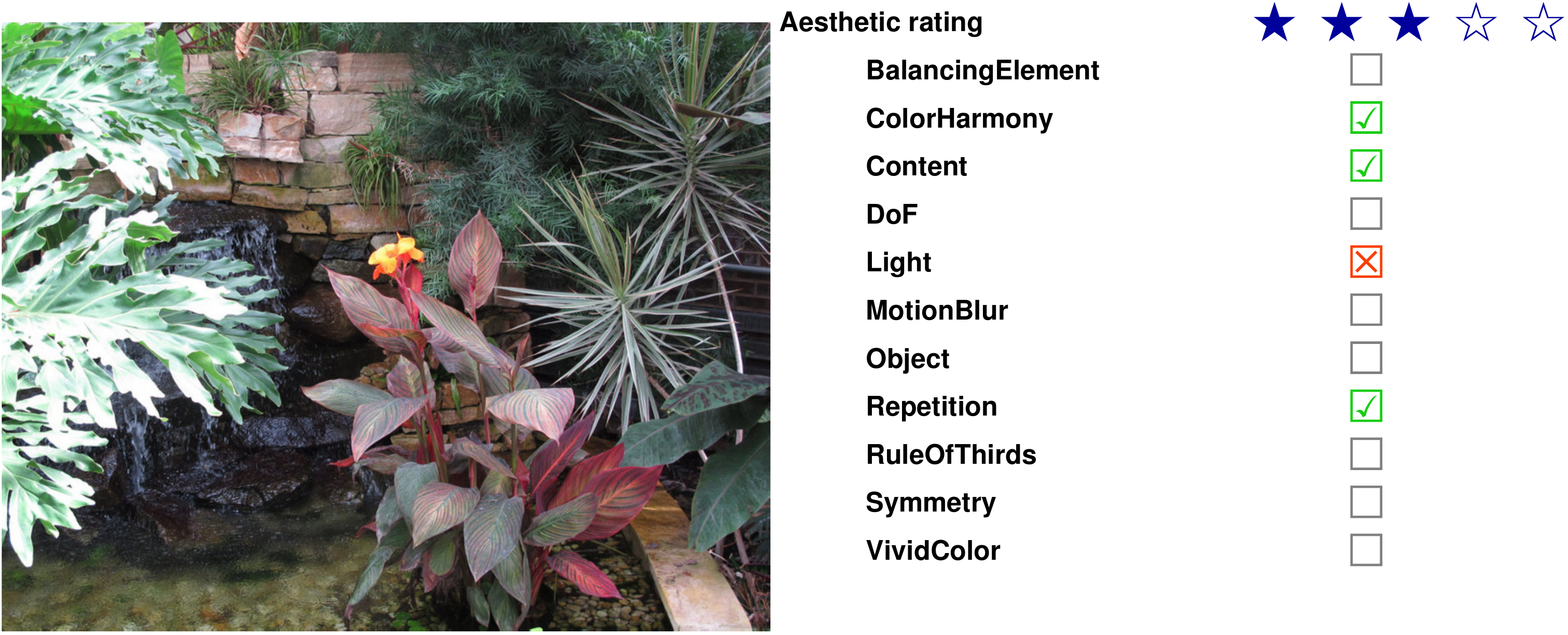}
    \end{minipage}
    \vspace{10mm}
    \caption{Some images outside our database with medium estimated scores.}
    \label{fig:mediumImages}
\end{figure*}

In this section,
we test our model on personal photos qualitatively,
in which these photos are downloaded online and not part of our AADB dataset.
As our model can predicts all the eleven attributes,
we show the attributes' estimation as well as the rated aesthetic scores.
For better visualization,
we simple set thresholds as  $(-0.2)$ and $(0.2)$ to characterize ``negative'', ``null'' and  ``positive'' attributes, respectively.
Figure~\ref{fig:highImages} -- \ref{fig:mediumImages}  show the results for images with high, low and medium estimated scores.
We can see, in general, our model reasonably captures attributes and gives aesthetic scores.

\end{document}